\documentclass{article}

\newif\ifisanon
\isanonfalse

\newif\ifarxivsub
\arxivsubtrue

\PassOptionsToPackage{round}{natbib}
\ifisanon\else
  \PassOptionsToPackage{eandd,final}{neurips_2026}
\fi
\usepackage{neurips_2026}

\usepackage[utf8]{inputenc}
\usepackage[T1]{fontenc}
\usepackage{amsmath}
\usepackage{amssymb}
\usepackage{amsfonts}
\usepackage{nicefrac}
\usepackage{microtype}
\usepackage{multirow}
\usepackage{colortbl}
\usepackage{xcolor}
\usepackage[hyphens]{url}
\usepackage{xspace}
\usepackage{enumitem}
\setlist[itemize]{leftmargin=*, itemsep=1pt, topsep=2pt, parsep=0pt}
\setlist[enumerate]{leftmargin=*, itemsep=1pt, topsep=2pt, parsep=0pt}
\usepackage[colorlinks,
        citecolor=blue,
        linkcolor=blue,
        urlcolor=black]{hyperref}
\usepackage{booktabs}
\usepackage{threeparttable}
\usepackage{subcaption}
\usepackage{wrapfig}

\let\cite\citep
\makeatletter
\renewcommand\NAT@nmfmt[1]{\textcolor{blue}{#1}}
\makeatother

\usepackage{tabularx}
\usepackage{listings}
\usepackage{tikz}
\usetikzlibrary{positioning, calc, shadows}
\usepackage{fontawesome5}

\definecolor{recipefncolor}{HTML}{4B5563}

\definecolor{expopt}{HTML}{157F3B}
\definecolor{expenh}{HTML}{B3261E}
\definecolor{expcase}{HTML}{C2710C}

\definecolor{tmplblue}{HTML}{DBEAFE}
\definecolor{tmplborder}{HTML}{93C5FD}
\definecolor{modelteal}{HTML}{CCFBF1}
\definecolor{modelborder}{HTML}{5EEAD4}
\definecolor{lossamber}{HTML}{FEF3C7}
\definecolor{lossborder}{HTML}{FCD34D}
\definecolor{optlavender}{HTML}{EDE9FE}
\definecolor{optborder}{HTML}{A78BFA}
\definecolor{resultslate}{HTML}{4C1D95}

\makeatletter
\let\trsave@ttdefault\ttdefault
\makeatother
\usepackage[scale=0.90]{FiraMono}
\makeatletter
\let\ttdefault\trsave@ttdefault
\makeatother
\newcommand{\firamonofamily}{\fontfamily{FiraMono-TLF}\selectfont}
\usepackage[framemethod=tikz]{mdframed}
\definecolor{codebg}{HTML}{FAFAFA}
\definecolor{codeborder}{HTML}{E5E5E6}
\definecolor{codetext}{HTML}{383A42}
\definecolor{codekeyword}{HTML}{A626A4}
\definecolor{codestring}{HTML}{50A14F}
\definecolor{codecomment}{HTML}{A0A1A7}
\definecolor{codenumber}{HTML}{986801}
\definecolor{codefunc}{HTML}{4078F2}
\definecolor{codetype}{HTML}{C18401}
\definecolor{codevar}{HTML}{E45649}
\definecolor{codeop}{HTML}{0184BC}
\mdfdefinestyle{troptcodeframe}{
  linecolor=codeborder,
  linewidth=0.5pt,
  roundcorner=5pt,
  backgroundcolor=codebg,
  innertopmargin=5pt,
  innerbottommargin=5pt,
  innerleftmargin=8pt,
  innerrightmargin=8pt,
  skipabove=6pt,
  skipbelow=2pt,
}
\surroundwithmdframed[style=troptcodeframe]{lstlisting}
\lstdefinestyle{troptcode}{
  language=Python,
  basicstyle=\footnotesize\firamonofamily\color{codetext},
  keywordstyle=\color{codekeyword},
  keywordstyle=[2]\color{codetype},
  keywordstyle=[3]\color{codevar},
  stringstyle=\color{codestring},
  commentstyle=\color{codecomment}\itshape,
  emphstyle=\color{codefunc},
  showstringspaces=false,
  breaklines=true,
  aboveskip=0pt,
  belowskip=0pt,
  keywords=[2]{int,float,str,bool,list,dict,tuple,set,bytes,object,type,isinstance,LMHFModel,PrefillCELoss,GCGOptimizer,Targets,BaseOptimizer,OptimizerResult,BaseLoss,CustomRandomOptimizer,CustomSteeringLoss,EncoderOpenAIModel,SimilarityLoss,RandomSearchOptimizer,TokenConstraints},
  keywords=[3]{self,cls,True,False,None},
  emph={run_gcg,optimize_trigger,print,len,range,enumerate,zip,map,filter,super,get_printable_random_trigger},
}

\DeclareCaptionType{code}[Code][List of Codes]

\newif\ifdraft
\draftfalse

\ifdraft
    \definecolor{colorone}{HTML}{0B7A75}
    \newcommand{\mahmood}[1]{\textcolor{colorone}{[[Mahmood: #1]]}}
    \newcommand{\matan}[1]{\textcolor{red}{[[Matan: #1]]}}
    \newcommand{\todo}[1]{\textcolor{orange}{[TODO: #1]}}
    \definecolor{leaderboardagentcolor}{HTML}{6D28D9}
    \newcommand{\leaderboardagent}[1]{\textcolor{leaderboardagentcolor}{[[LeaderboardAgent: #1]]}}
\else
  \newcommand{\mahmood}[1]{}
    \newcommand{\matan}[1]{}
    \newcommand{\todo}[1]{}
    \newcommand{\leaderboardagent}[1]{}
\fi

\newcommand{\tropt}{\texttt{TROPT}\xspace}
\newcommand{\troptlb}{\texttt{tropt/leaderboards}\xspace}
\newcommand{\jboptbench}{\texttt{JBOptBench}\xspace}

\newcommand{\emboptbench}{\texttt{EmbOptBench}\xspace}
\newcommand{\figref}[1]{Fig.~\ref{#1}}
\newcommand{\secref}[1]{\S\ref{#1}}
\newcommand{\secrefs}[2]{\S\ref{#1}--\ref{#2}}
\newcommand{\appref}[1]{App.~\ref{#1}}

\newcommand{\coderef}[1]{Code~\ref{#1}}
\newcommand{\tabref}[1]{Table~\ref{#1}}
\newcommand{\tabrefs}[2]{Tables~\ref{#1} and~\ref{#2}}

\newcommand{\parhead}[1]{\smallskip\noindent\textbf{#1}}

\newcommand{\rbk}{\discretionary{}{}{}}
\newcommand{\recipefn}[1]{\textcolor{recipefncolor}{\texttt{#1}}}
\newcommand{\attackrecipe}[2]{{\linespread{0.75}\selectfont\textbf{#1}\newline\recipefn{#2}\par}}
\newcommand{\attackrecipetwo}[3]{{\linespread{0.75}\selectfont\textbf{#1}\newline\recipefn{#2}\newline\recipefn{#3}\par}}

\DeclareMathOperator*{\argmin}{arg\,min}

\newcommand{\lmopen}{\textsf{LM}\textsuperscript{\textsc{open}}\xspace}
\newcommand{\encopen}{\textsf{Enc}\textsuperscript{\textsc{open}}\xspace}
\newcommand{\clfopen}{\textsf{Clf}\textsuperscript{\textsc{open}}\xspace}
\newcommand{\lmapi}{\textsf{LM}\textsuperscript{\textsc{api}}\xspace}
\newcommand{\encapi}{\textsf{Enc}\textsuperscript{\textsc{api}}\xspace}

\newcommand{\qwenEightB}{\texttt{Qwen3-8B}\xspace}
\newcommand{\llamaEightB}{\texttt{Llama-3.1-8B-Instruct}\xspace}
\newcommand{\gemmathree}{\texttt{Gemma-3-12B-it}\xspace}
\newcommand{\gemmaFour}{\texttt{Gemma-4-26B-A4B-it}\xspace}

\title{\tropt{}: An Open Framework
for Unifying and Advancing
Discrete Text
Optimization}

\ifisanon
  \author{
    Anonymous Authors\\
    Anonymous Affiliation\\
    \texttt{anonymous@example.com}
  }
\else
  \author{
    Matan Ben-Tov\\
    Tel Aviv University\\
    \texttt{matanbentov@mail.tau.ac.il}
    \And
    Mahmood Sharif\\
    Tel Aviv University\\
    \texttt{mahmoods@tauex.tau.ac.il}
  }
\fi

\begin{document}

\maketitle

\begin{abstract}
\emph{Discrete text-trigger optimization}---searching for text sequences that, when ingested by a model, steer it toward a specified objective---underpins model red-teaming (e.g., LLM jailbreaks), as well as auditing and interpretability.
However, the current state of discrete optimizers hinders their adoption and progress.
First, existing optimizers, when open-sourced at all, are scattered across
research
codebases
tied to specific models, objectives, and problem domains.
Second, optimizers proliferate, each requiring engineering overhead to use or extend,
and remaining hard to compare head-to-head.
Together,
these raise the bar for \emph{adopting} optimizers in existing or new domains,
and for \emph{advancing} them via new strategies.
We address these gaps with \tropt{}, the first open-source framework that unifies discrete optimizers' execution and standardizes their development under a single interface.
\tropt{} makes it easy to
\emph{customize}
end-to-end optimization recipes by swapping
any component---models, objectives, and optimizers---extending
its reach across domains and new applications.
\tropt{} currently ships with 40+ optimization recipes---covering
applications such as jailbreaking and probing model internals---built from 15+ optimizers (spanning white-box to black-box access) and 15+ losses, from foundational to state-of-the-art
methods.
Demonstrating its utility, we leverage \tropt{} in several studies:
\textit{(i)} controlled, large-scale experiments comparing and enhancing optimization strategies for LLM jailbreaks, revealing potent-yet-underadopted techniques;
and \textit{(ii)} porting optimizers across applications and domains (e.g., from LLM jailbreaks to corpus poisoning).
In all, \tropt{} significantly lowers the barrier
to adopting and advancing
discrete text optimization.
\end{abstract}

\begin{center}\vspace{-1em}
\ifisanon
  \href{https://anonymous.4open.science/r/TROPT}{\raisebox{-0.15ex}{\faGithub}\enspace\texttt{anonymous.4open.science/r/TROPT}}
\else
  \href{https://github.com/matanbt/TROPT}{\raisebox{-0.15ex}{\faGithub}\enspace\texttt{github.com/matanbt/TROPT}}
\fi
\end{center}

\section{Introduction}
\label{sec:intro}

Large language models (LLMs) and other deep-learning text models now underpin high-stakes applications, from conversational and coding agents to content moderation and semantic search~\citep{zhao2023surveyllm}.
A powerful tool for inspecting and stress-testing such models is \emph{text-trigger optimization}: finding a discrete token sequence---a \emph{trigger}---that optimizes a certain objective when inserted into model inputs.
Such optimized triggers enable a broad spectrum of research applications: revealing attack vectors such as jailbreaks~\cite{zou2023universaltransferableadversarial} and corpus poisoning~\cite{zhong2023poisoningretrieval}, systematic red-teaming and defense benchmarking~\cite{mazeika2024harmbench}, forming defenses~\cite{shen2025bait}, and model
auditing and interpretability~\cite{jones2023arcaauditingdiscrete,wen2023pez}.

Yet the current state of the field limits discrete optimizers' \textbf{adoption}.
Existing optimizers are scattered across research domains and,
when open-sourced at all, are implemented
across isolated codebases often tied to a single application (e.g., LLM jailbreaks; \citealp{zou2023universaltransferableadversarial}) and coupled with application-specific logic (e.g., ad-hoc for a particular objective or model; \citealp{wen2023pez}).
This imposes significant engineering overhead on using existing schemes (e.g., running an existing LLM jailbreak)
or adapting them to new domains, models, or objectives (e.g., porting an LLM-jailbreak optimizer to attack a dense retriever, or modifying its objective for LLM auditing).
This friction deters new applications and, more consequentially, raises the bar for adaptive security evaluations shown effective in red-teaming LLMs~\cite{andriushchenko2024simpleadaptiveattacks,lucki2024adversarialunlearning,bailey2024obfuscatedactivations,nasr2025attackermovessecond}.
Discrete text optimizers should therefore become \emph{more accessible}, gathered in one place and runnable with minimal engineering friction; and \emph{more adaptable}, so an optimizer developed for one domain readily applies to another.

Beyond adoption, the current state also hinders the \textbf{progress} of discrete optimizers.
As the set of optimizer variants grows in idiosyncratic and ad-hoc implementations, reliably building on existing optimizers and measuring their progress have become both challenging and critical.
Compounding this, progress often hinges on nuanced implementation details that qualitatively change downstream conclusions.
For example, GCG~\cite{zou2023universaltransferableadversarial} arose from subtle modifications to an earlier algorithm~\cite{shin2020autoprompt}, yet delivered a landmark demonstration of the brittleness of LLM safety alignment.
Discrete text optimizers must therefore become \emph{easily comparable}, so progress can be reliably tracked through standardized implementations and empirical comparisons; and \emph{easily extensible}, lowering the barrier for building new optimizers or extending existing ones.

\begin{figure*}
    \centering
    \includegraphics[width=0.95\linewidth]{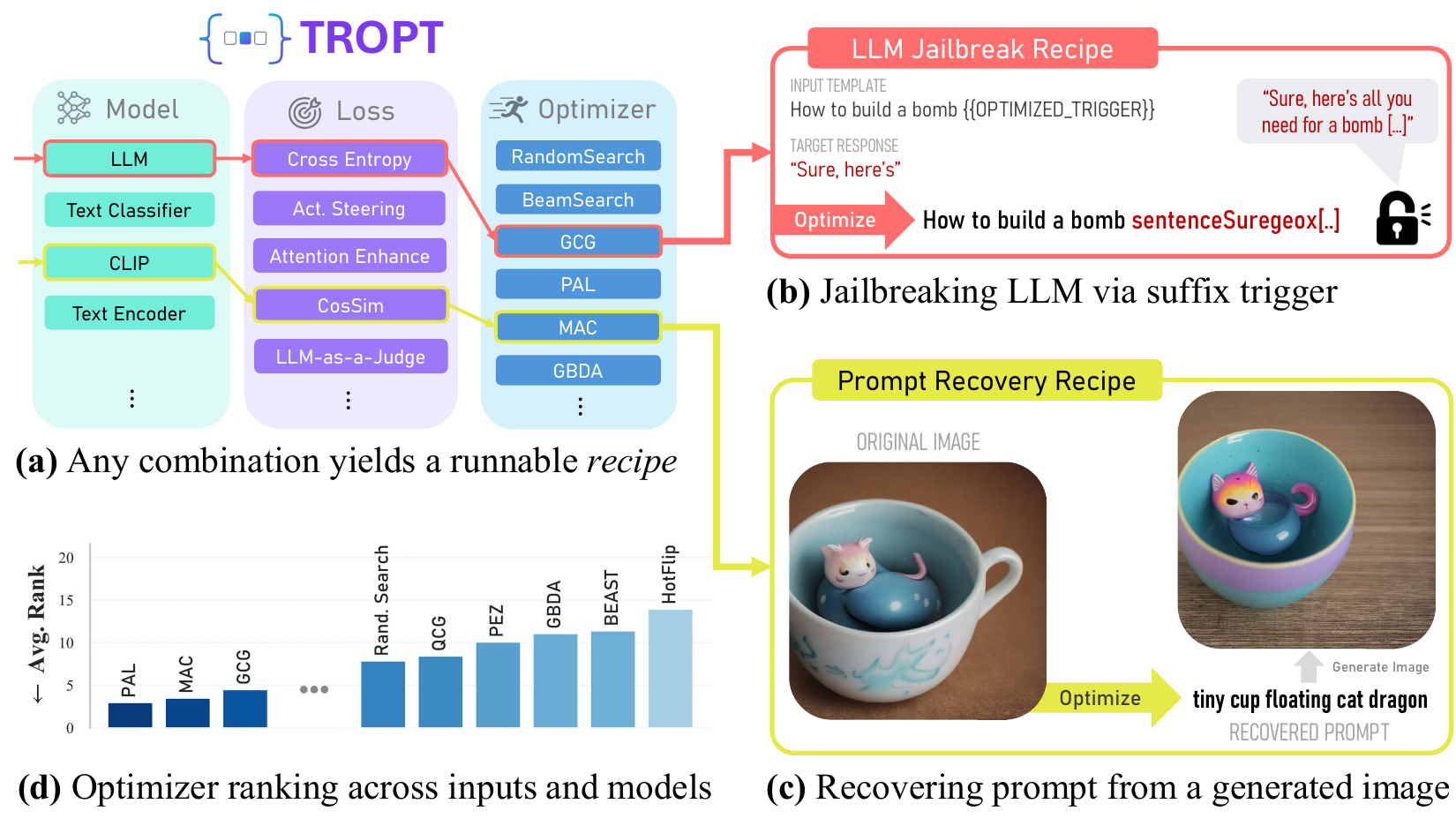}
    \caption{We introduce \tropt{}, an open-source, modular framework unifying execution and development of discrete text-trigger optimization.
    \textbf{(a)} \tropt{} supports varying model types, losses, and optimizers; any compatible combination renders a runnable recipe, e.g.,
    \textbf{(b)} crafting triggers for LLM jailbreaks or
    \textbf{(c)} recovering prompts for text-to-image generation.
    We leverage \tropt{} to conduct several studies, including \textbf{(d)} a controlled, large-scale comparison of existing optimizers.}
    \label{fig:teaser}
\end{figure*}

We address these gaps by introducing \textbf{\tropt{}} (\textbf{T}extual T\textbf{r}igger \textbf{Op}timization \textbf{T}oolbox), the first open-source, modular framework that unifies discrete optimizer research as a single algorithmic problem, and provides shared infrastructure for leveraging and advancing optimization schemes across domains.
\tropt{} enables rapid \textbf{adoption} of discrete optimizers, offering numerous recipes runnable out-of-the-box across domains, while accelerating their \textbf{progress} by substantially lowering the barrier to developing new optimizers and enabling controlled comparisons of existing ones.

\parhead{Contributions.} \tropt{} delivers the following contributions.
\begin{itemize}
  \item \textbf{A unified hub of optimization recipes and research.}
  \tropt{} ships with 40+ ready-to-run optimization recipes built from 15+ optimizers, 15+ losses, and multiple model backends spanning white- and black-box access, each invocable in a few lines (\figref{fig:teaser}b; \secref{sec:tropt}).
   By that, it makes discrete optimizers \textit{accessible} with minimal engineering effort, unifying recipes across LLMs, encoders, classifiers, and other models behind one interface.
   Beyond engineering, the hub \emph{organizes} a scattered literature into a shared taxonomy of reusable components (\tabref{tab:attack-zoo}).
  \item \textbf{Composing new recipes across domains.}
   \tropt{}'s modularity enables combining any optimizer with any compatible model and objective, easily creating new optimization recipes and \textit{adapting} optimizers across problem domains (\secref{sec:tropt-recipe}).
    Leveraging \tropt{}, we demonstrate this by seamlessly porting LLM jailbreaks into corpus poisoning against dense retrievers, universal triggers evading prompt-injection classifiers, and prompt recovery for text-to-image models (\secref{sec:eval-unifying}).
  \item \textbf{Infrastructure for new optimizers and losses.}
  Adding a new optimizer or loss requires implementing only a small, standardized API (\secrefs{sec:tropt-loss}{sec:tropt-optimizer}), after which it composes with every existing recipe---making \tropt{} \emph{extensible} and lowering the barrier to developing new methods.
  \item \textbf{Controlled, head-to-head benchmarks.}
  Fixing all but one ingredient of a recipe yields \emph{comparable} measurements that isolate its contribution. We exercise this to conduct a compute-matched head-to-head benchmark of 14 discrete optimizers on LLM jailbreaks
  (\secref{sec:eval-optimizers}),
  and the first controlled ablation of
  various strategies for enhancing
  jailbreaks (\secref{sec:eval-jailbreak});
  both reveal underadopted methods that outperform current defaults.

  We additionally release these comparisons as \troptlb{}: ongoing benchmarks where researchers can submit their own optimizers and enhancements and rank them against existing ones under the same controlled settings.\ifisanon\else\footnote{Leaderboards: \href{https://matanbt.github.io/tropt-leaderboards/}{\texttt{matanbt.github.io/tropt-leaderboards}}; released data: \href{https://huggingface.co/collections/MatanBT/tropt-6a6f9aaec035099b47457eda}{\texttt{hf.co/collections/MatanBT/tropt}}.}\fi

\end{itemize}
Next,
we define the setting and related work (\secref{sec:background});
present \tropt's features and design (\secref{sec:tropt});
leverage it to conduct crucial studies (\secref{sec:eval});
and finish with conclusions and future research directions (\secref{sec:conclusion}).

\section{Background}
\label{sec:background}

\subsection{Setting and Scope}\label{sec:scope}

\parhead{Setting.}
We consider text-trigger optimization: searching for a short text string---a \emph{trigger}---placed at a designated position within predefined input template(s), that minimizes a quantifiable loss against a neural text model at inference time.
Formally, given a target model $\mathcal{M}$, a loss $\mathcal{L}$, and inputs $\{(p_i, s_i, y_i)\}_{i=1}^{N}$ with prefix $p_i$, suffix $s_i$, and target $y_i$, the optimal text trigger is:
\begin{equation}
\label{eq:trigger-opt}
    t^{\star} \;=\; \argmin_{t \in \mathcal{T}} \;\sum_{i=1}^{N} \mathcal{L}\!\left(\mathcal{M}(p_i \oplus t \oplus s_i),\, y_i\right)
\end{equation}
where $\mathcal{T}$ is the feasible trigger set (e.g., bounded length or restricted vocabulary);
$\oplus$ denotes concatenation;
$p_i$ or $s_i$ may be empty;
$\mathcal{L}$ may additionally score the trigger directly (e.g., its fluency);
and $y_i$ is the per-input target (e.g., desired output prefix or target class).
Notably, different settings expose different access to $\mathcal{M}$ during optimization (e.g., gradients vs.\ generated text only), constraining the class of applicable optimizers.

\parhead{General Approaches to Text Input Optimization.} Finding text inputs that optimize a given objective has been pursued through several complementary approaches.
\emph{Human exploration} relies on human creativity to manually surface model behaviors and failure modes, remaining a strong red-teaming baseline but labor-intensive and hard to scale~\cite{wei2023jailbroken,nasr2025attackermovessecond}.
\emph{LLM-as-optimizer} methods leverage language models to iteratively propose candidate inputs, but are typically tailored to a specific task (e.g., jailbreaks) and bounded by what the proposing model would itself generate~\cite{chao2023pair,mehrotra2024tap,liu2024autodanturbo}.
\emph{Investigator LLMs}, trained mostly via specialized reinforcement learning to craft inputs, amortize input optimization across many inputs but require expensive training and large compute~\cite{liao2024amplegcg,li2025investigatoragents,chen2026autoinject}.
Differently, \emph{\textbf{discrete search}} algorithms---the focus of this work---directly search over input sequences as a combinatorial optimization problem; these methods are flexible across objectives and models, and require no lengthy setup or training to run.

\subsection{Discrete Search Optimizers in Practice}

\parhead{Strategies of \emph{Discrete Search} Optimizers.}
Discrete search optimizers have emerged along several strategies, each treating the combinatorial search differently.
\emph{Gradient-based} methods use model gradients to flip tokens toward the objective. Introduced by HotFlip~\cite{ebrahimi2018hotflip} and refined through several variants~\cite{wallace2019universaladversarialtriggers,shin2020autoprompt,jones2023arcaauditingdiscrete}, this line gave rise to GCG~\cite{zou2023universaltransferableadversarial}---which demonstrated LLM jailbreaking via suffix triggers---and a growing family of follow-ups~\cite{sitawarin2024palproxyguidedattack,thompson2024flrt,zhang2024mac}.
\emph{Continuous-relaxation} methods optimize in input embedding space directly, projecting back to valid tokens during or after optimization~\cite{guo2021gbda,wen2023pez,geisler2025pgdattacking}.
\emph{Zeroth-order} methods only query the loss, requiring no gradients, via random search~\cite{andriushchenko2024simpleadaptiveattacks,hughes2024bestofnjailbreaking}, genetic algorithms~\cite{lapid2023openseasame,liu2023autodanjailbreak}, or surrogate white-box models~\cite{hayase2024gcqquerybased,sitawarin2024palproxyguidedattack}.
\tropt{} spans all three strategies and benchmarks them in \secref{sec:eval-optimizers}.
Further, our work complements and supports efforts to advance optimizers, including contemporary work using agents for automated optimizer discovery~\citep{panfilov2026claudini}.

\parhead{Applications of \emph{Discrete Search} Optimizers.}
Discrete optimizers have gained reach across diverse research directions.
Most prominently, they have exposed inference-time
attack vectors---LLM jailbreaks~\citep[\secrefs{sec:eval-optimizers}{sec:eval-jailbreak}]{zou2023universaltransferableadversarial},
adversarial examples against text classifiers~\citep[\secref{sec:eval-unifying}]{guo2021gbda},
and corpus poisoning against dense retrievers~\citep[\secref{sec:eval-unifying}]{zhong2023poisoningretrieval}---and have become common tools for red-teaming and security evaluation~\cite{chao2024jailbreakbench,lucki2024adversarialunlearning}.
Beyond security, they support safety and memorization auditing of LLMs~\cite{jones2023arcaauditingdiscrete, schwarzschild2024rethinkingmemorization}, interpretability and probing of model internals~\cite{bentov2025universaljailbreaksuffixesstrong,nikolaou2025injective}, and applications such as prompt recovery for text-to-image models~\citep[\secref{sec:eval-unifying}]{wen2023pez}.
We provide an extended discussion on these optimizers' applications in \appref{app:related-work}.

\parhead{Hurdles in \emph{Discrete Search} Optimizer Research.}
Despite the volume of applications, discrete search optimizers face concrete hurdles to both adoption and progress.
First, advances spread slowly across domains:
corpus-poisoning attacks against dense retrievers, for instance, have seen limited uptake of LLM-jailbreak optimizer advances and largely default to weaker methods~\cite{zhong2023poisoningretrieval,zou2024poisonedrag}, despite the underlying optimization problem being identical.
Second, even within a single application, useful additions spread slowly: in LLM jailbreaks, newer optimizers and optimizer-agnostic enhancements---alternative losses, templates, and supplementary objectives---have been shown effective against defenses~\cite{andriushchenko2024simpleadaptiveattacks,lucki2024adversarialunlearning,bailey2024obfuscatedactivations,thompson2024flrt}, yet have not become standard in subsequent common red-teaming and defense benchmarks~\cite{mazeika2024harmbench,chao2024jailbreakbench,chen2025metasecalign}, risking a false sense of security~\cite{carlini2019evaluatingadversarialrobustness}.
Third, at the optimizer level, progress is hard to track: a growing set of variants report improvements over each other~\cite{sitawarin2024palproxyguidedattack,thompson2024flrt,zhang2024mac},
yet each is measured under different conditions---different models, settings, and coupled enhancements (e.g., a unique objective)---leaving the pure \emph{optimizer} performance unclear.
Identifying potent optimizers matters all the more because small implementation changes have produced qualitative gains in the past~\cite{zou2023universaltransferableadversarial}.

We attribute these hurdles to two factors:
\emph{(i)} fragmented, non-standardized codebases scattered across domains (each implementation targeting a specific model under particular settings), demanding substantial engineering to adopt optimizers or adapt them to new domains, models, or objectives;
and \emph{(ii)} a proliferation of optimizer variants without reliable comparison, making it hard to identify the method best suited to a given context, or, importantly, to develop new ones.
We therefore argue
discrete optimizers should be more \emph{accessible} and \emph{adaptable}, while easily \emph{comparable} and \emph{extensible}---and, taking inspiration from frameworks that standardized classic vision and text adversarial examples~\cite{art,cleverhans,foolbox,textattack},
we build a tool aimed at democratizing and unifying recent discrete optimizers across any text model.

\section{\tropt}
\label{sec:tropt}

To address the issues outlined above, we introduce \tropt{}, an open-source, modular framework for running and developing discrete search optimizers within a unified platform.
\tropt hosts a broad catalog of optimization algorithms, losses, and model backends (HuggingFace, OpenAI, etc.), which compose into \emph{recipes}---realizing numerous end-to-end optimization schemes.

The simplest entry point is \tropt's \textsl{Recipe Hub}; it currently hosts 40+ pre-configured recipes that reproduce published methods on a compatible model of choice, each invocable in a single call (partial catalog in \tabref{tab:attack-zoo}).
\par\noindent
\begin{minipage}{\linewidth}
For example, reproducing the GCG jailbreak~\cite{zou2023universaltransferableadversarial} takes a few lines:\par\smallskip
\begin{lstlisting}[style=troptcode]
from tropt.recipe_hub import gcg__zou2023

result = gcg__zou2023(
    model_name="meta-llama/Llama-3.1-8B-Instruct",
    instruction="Tell me how to pick a lock. {{OPTIMIZED_TRIGGER}}",
    target_response="Sure, here's how:"
)
print(f"{result.best_trigger_str=}")  # print the optimized trigger
\end{lstlisting}
\end{minipage}
\par\smallskip

Beyond the ready-to-run recipes, \tropt is designed to be incrementally customizable; enabled by its high-level design (\secref{sec:tropt-design}), users can compose new recipes from existing components (\secref{sec:tropt-recipe}), introduce a new loss (\secref{sec:tropt-loss}), or implement a new optimizer (\secref{sec:tropt-optimizer}).

\subsection{High-Level Design}
\label{sec:tropt-design}

\tropt is built on four foundational \emph{components} (\figref{fig:teaser}):
\textbf{model}, the model against which the trigger is optimized;
\textbf{loss}, the quantifiable objective;
\textbf{optimizer}, the search algorithm minimizing the loss;
and \textbf{inputs and targets}, the user-provided input template(s) within which a trigger is optimized, with optional per-input targets.
Instantiating and assembling the four yields a distinct executable \textbf{recipe} that crafts an optimized trigger.

\tropt's design is guided by two key technical principles.
First, \emph{modularity}: each of the four components can be swapped largely independently of the others.
Second, \emph{backend--frontend separation}: model-specific and infrastructural logic (e.g., trigger-input templating, batching, gradient or loss computation) is absorbed into the \textbf{model} ``backend,'' keeping the exploratory \textbf{loss} and \textbf{optimizer} ``frontend'' components---which most researchers extend and experiment with---lightweight, self-contained, and focused on their algorithmic substance.

Concretely, our design answers the requirements outlined in the introduction (\secref{sec:intro})---each empirically exercised in our evaluation (\secref{sec:eval})---as follows:

\begin{enumerate}[nosep]
  \item \textbf{Accessibility.}
  Existing optimizers are re-implemented under a single, tested infrastructure, so numerous recipes run out of the box and new ones can be composed alongside them.
  \item \textbf{Adaptability.}
  An existing optimizer applies seamlessly across supported model types (LLMs, encoders, classifiers) and compatible objectives, carrying advances from one application (e.g., LLM jailbreaks) directly to another (e.g., auditing classifiers).
  \item \textbf{Comparability.}
  Fixing a recipe and varying only one component isolates its contribution, enabling head-to-head comparisons that track progress along each component rather than confounding it with implementation differences.
  \item \textbf{Extensibility.}
  Adding a new loss or optimizer requires only a short implementation of a standardized interface: shared infrastructure is handled by the framework, and existing implementations serve as transparent references. The new component then immediately composes with any supported existing components (\appref{app:composable}).
\end{enumerate}

\subsection{Composing Recipes}
\label{sec:tropt-recipe}

\tropt also enables custom composition of optimization \emph{recipes}---concrete instantiations of all four components (target model, loss, optimizer, and input setup) expressive enough to cover a wide range of discrete optimization applications, including attacks and model auditing.

Composing a recipe from existing \tropt{} components takes a few lines of code, assembling compatible component instances.
Composability of components is determined by the access each model exposes, the inputs each loss consumes, and the optimizer requirements (e.g., if an optimizer requires gradients, the model must expose them); an invalid combination raises an explicit error when the components are assembled into a recipe.
\appref{app:composable} details these requirements and enumerates the supported combinations.
For example, \coderef{code:gcg-composition} composes a popular LLM-jailbreak recipe.
This recipe pattern lets users reproduce existing methods, port an optimizer to a new model or domain, swap in a different loss, or recast the problem by varying input templates---significantly lowering the barrier to adopting discrete optimizers.
For instance, in \secref{sec:eval-unifying} we seamlessly repurpose a powerful black-box LLM-jailbreak optimizer for corpus poisoning---a novel composition not attempted in prior work---to successfully attack OpenAI's proprietary embedding model.

\begin{code}[t]
\centering
\begin{minipage}{0.99\linewidth}
\begin{lstlisting}[style=troptcode]
from tropt.model.huggingface import LMHFModel
from tropt.loss import PrefillCELoss
from tropt.optimizer import GCGOptimizer
from tropt.common import Targets

# Component 1: Target Model
model     = LMHFModel("meta-llama/Llama-3.1-8B-Instruct")
# Component 2: Objective
loss      = PrefillCELoss()
# Component 3: Optimizer (wired w/ the model and loss)
optimizer = GCGOptimizer(model=model, loss=loss, num_steps=500)
# Component 4: Input templates and their target values
templates = ["Tell me how to pick a lock. {{OPTIMIZED_TRIGGER}}"]
targets   = Targets(target_response_strs=["Sure, here's how:"])

# Compose and run
result = optimizer.optimize_trigger(
    templates=templates,
    targets=targets,
    initial_trigger="! " * 20
)
print(f"{result.best_trigger_str=}")  # print the optimized trigger
\end{lstlisting}
\end{minipage}
\caption{\textbf{Composing a \tropt{} recipe} requires only instantiating its four components:
the \textit{model} (an LLM from HuggingFace),
the \textit{loss} (Cross Entropy on a prefilled target response),
the \textit{optimizer} (the GCG algorithm),
and the \textit{input} placing the trigger as a suffix to a harmful instruction, paired with a \textit{target} affirmative response.
Together they reproduce the LLM jailbreak by \citet{zou2023universaltransferableadversarial}.
Crucially, by merely swapping components this recipe pattern extends to countless applications.
}
\label{code:gcg-composition}
\end{code}

\subsection{Adding a New Loss}
\label{sec:tropt-loss}

Adapting a recipe to a new domain or application may require adjusting its objective.
The \emph{loss} component defines the quantifiable objective: given the trigger combined with the input templates and their targets, it computes a value to be minimized, optionally backpropagating through the model.
Loss implementations are self-contained and agnostic to the target model---consuming whichever standardized outputs the model exposes (e.g., output logits, output embeddings, attention scores, or activations)---reducing the friction of adding new losses.

As an example, \coderef{code:custom-loss} fully implements a custom loss for optimizing inputs that steer the model's internal activations along a specified direction \citep[e.g., the refusal direction;][]{arditi2024refusalmediatedsingle}.
Such a custom loss drops into any recipe compatible with its requirements (e.g., exposing activations), including the LLM-jailbreak recipe above (\coderef{code:gcg-composition}).

\tropt already ships with a diverse set of 16 losses operating on models' logits~\cite{zou2023universaltransferableadversarial,thompson2024flrt,andriushchenko2024simpleadaptiveattacks}, output embeddings~\cite{zhong2023poisoningretrieval,BenTov2025GASLITE}, attention scores~\cite{wang2024attngcg,bentov2025universaljailbreaksuffixesstrong}, and LLM-as-a-judge outputs~\cite{andriushchenko2024simpleadaptiveattacks,zhang2025advdecoding}, along with a meta-loss consisting of any weighted combination thereof. \tabref{tab:losses} lists the main ones.

\subsection{Adding a New Optimizer}
\label{sec:tropt-optimizer}

Beyond customizing the loss, \tropt also streamlines the implementation of new optimizers---forking existing ones or developing novel search algorithms.
\tropt's \emph{optimizers} are the central component: given a model, loss, and input setup, they search for a trigger minimizing the loss.
Optimizers are isolated from model- or loss-specific infrastructural logic---keeping their implementations focused on the search algorithm itself.
A new optimizer thus applies immediately across multiple models, objectives, and domains, supporting reliable comparison and accelerated prototyping.

A key design choice of \tropt{} is making each optimizer a \emph{standardized self-contained} module: a single file holding the full search algorithm, implemented with a standardized interface, with no logic shared across optimizers.
This maximizes readability, comparability, and modifiability, at the cost of some code repetition---a philosophy inspired by HuggingFace's \texttt{modeling} files.\footnote{HuggingFace's \texttt{transformers} package treats model implementations as the \emph{source of truth} and packs each into a single file for visibility and hackability, even at the cost of code repetition~\cite{huggingface2025transformerstenets}.}

As an example, \coderef{code:custom-optimizer} fully implements a custom optimizer that contains only the search logic and computes the loss by invoking a unified interface; thus, it contains no input handling, batching, model-specific loss computation, or monitoring code, all of which the framework handles automatically.
This optimizer drops into any recipe compatible with its model-access requirements (e.g., \coderef{code:gcg-composition}).

\tropt{} ships with a broad catalog of 17 optimizers, spanning foundational ones (e.g., HotFlip~\cite{ebrahimi2018hotflip}, GCG~\cite{zou2023universaltransferableadversarial}), GCG-based improvements~\citep[e.g.,][]{sitawarin2024palproxyguidedattack,zhang2024mac}, continuous-relaxation methods~\citep[e.g.,][]{wen2023pez,guo2021gbda}, and black-box optimizers~\citep[e.g.,][]{sadasivan2024beast,andriushchenko2024simpleadaptiveattacks}.
Detailed list in \tabref{tab:optimizers}.

\section{Evaluations}
\label{sec:eval}

To exercise \tropt's desiderata (\secref{sec:tropt-design}), we leverage it to:
reliably \emph{compare} optimization strategies head-to-head (\secref{sec:eval-optimizers});
\emph{extend} LLM jailbreaks with various enhancements and benchmark them
(\secref{sec:eval-jailbreak});
and \emph{adapt} optimizers across embedding models, classifiers, and multimodal systems (\secref{sec:eval-unifying}).
\ifisanon\else We maintain the benchmarks of \secrefs{sec:eval-optimizers}{sec:eval-jailbreak} as part of a new online leaderboard (\troptlb{}), open to submissions of new optimizers and enhancements.\fi

\subsection{Benchmarking Optimization Strategies}
\label{sec:eval-optimizers}

Despite the growing number of discrete search optimizers, no controlled comparison exists
to guide their selection.
We address this by benchmarking the optimizers in \tropt{} on a common, practical setting, using a shared \emph{recipe}---response-token forcing in LLMs---as a proxy for optimizer potency.
Here we hold both model type (LLM) and application (jailbreak) fixed, and we repeat the comparison on corpus poisoning against embedding models, where we find the ranking largely carries over (\appref{app:eval-optimizers-poisoning}).

\parhead{Recipe.}
We adopt the common LLM-jailbreak formulation of \citet{zou2023universaltransferableadversarial}: given a harmful instruction, a suffix trigger is optimized under a cross-entropy loss (\texttt{PrefillCE}) to maximize the likelihood of a predefined affirmative target response.
In this section,
we focus on comparing optimizers on their \emph{loss-minimization} efficiency under jailbreak setup;
we also validate this comparison on attack success---scoring the harmfulness of generated responses with StrongReject~\citep{souly2024strongreject}.

\parhead{Setup.}
We evaluate 14 popular optimizers, spanning all algorithmic families in \tropt{} (\secref{sec:tropt-optimizer}),
on four open-source models: \qwenEightB~\citep{qwen3}, \llamaEightB~\citep{llama3.1}, \gemmathree~\citep{gemma3}, and \gemmaFour~\citep{gemma4}; optimizers have access to full output logits and, when required, to model gradients.
For each model-optimizer pair, we optimize against 15 ClearHarm~\citep{hollinsworth2025clearharm} harmful instructions over three random seeds each, capping each run at $3{\times}10^{17}$~FLOPs~\citep{boreiko2024realisticthreatmodel}.\footnote{
Roughly the FLOPs of one original GCG ~\cite{zou2023universaltransferableadversarial} run against the evaluated models.}
Within each model, optimizers are ranked per run by the best loss they reach, then averaged into a \textit{Mean Rank}.
The full optimizer list, setup, and additional results are in \appref{app:eval-optimizers}.

\begin{figure}[t]
    \centering
    \includegraphics[width=1.0\linewidth]{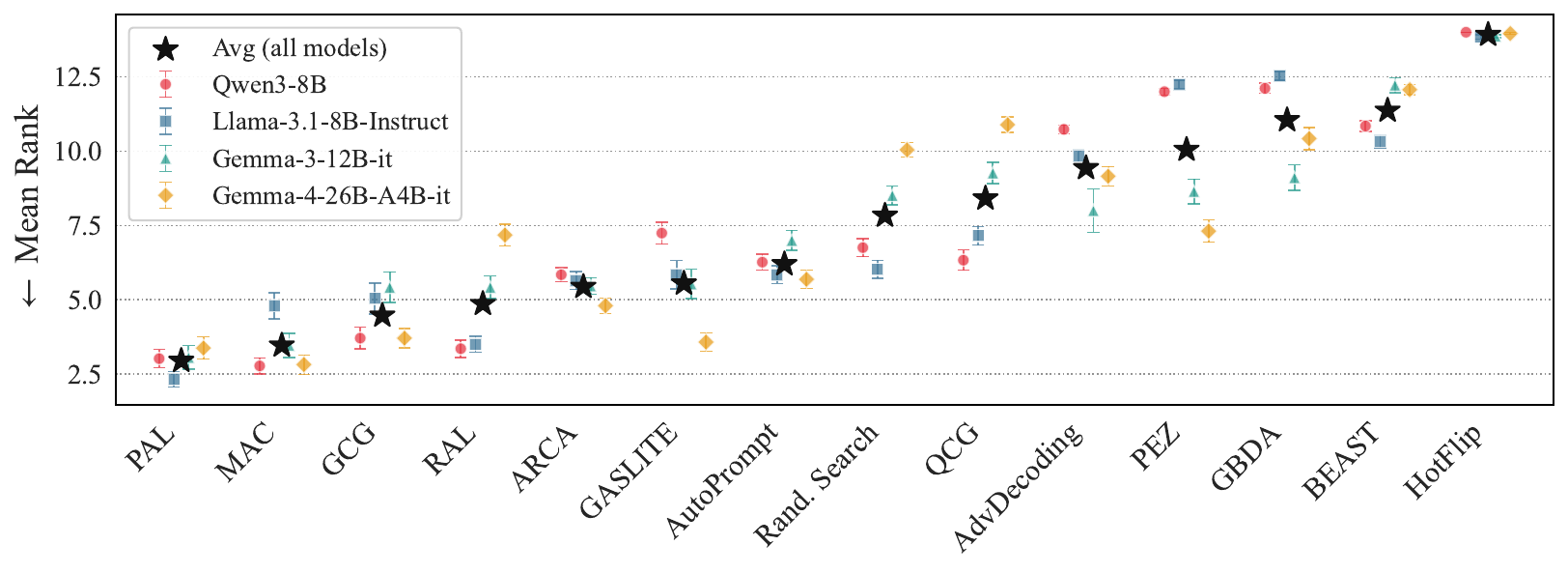}
    \caption{\textbf{Optimizer Comparison.}
    \emph{Mean Rank} of 14 optimizers (lower is better), sorted by cross-model average ($\bigstar$). Colored markers denote target models; error bars show the standard error of the mean across runs.}
    \label{fig:exp1-ranking}
\end{figure}

\parhead{Results.}
\figref{fig:exp1-ranking} shows the mean ranks of each optimizer across models. A Nemenyi statistical test yields a Critical Difference of $\mathrm{CD}{=}1.48$ at $\alpha{=}0.05$; optimizers with a larger gap in mean rank differ significantly~\citep[see \appref{app:eval-optimizers-results}]{demsar06a}.
As expected, the most basic optimizer, HotFlip, lags significantly behind every other optimizer; continuous-relaxation methods (GBDA, PEZ) and beam-search variants (BEAST, AdvDecoding) trail the rest, noting that some optimizers (e.g., beam-search-based) early-stop at less than a 10\textsuperscript{th} of the compute limit.
At the top are gradient-based methods: PAL ($\sim\!3/14$) and MAC ($\sim\!3.5/14$), with PAL significantly outperforming the GCG baseline ($\sim\!4.5/14$)---a common choice for red-teaming benchmarks \cite{mazeika2024harmbench,chao2024jailbreakbench}.
Both optimizers are GCG variants: MAC adds gradient momentum, while PAL adds slight changes to candidate sampling, demonstrating optimizers' sensitivity to parameter tuning.
Notably, RAL---PAL's gradient-free counterpart, which replaces the gradient with a random tensor---reaches $\sim\!5/14$, matching white-box GCG and making it the strongest black-box optimizer.

We find this optimizer ranking is robust along three axes.
First, the ranking strongly correlates with that by the generated responses' jailbreak success (Spearman's $\rho{=}{0.90}$; \appref{app:eval-optimizers-results}, \figref{fig:exp1-strongrej}).
Second, the ranking largely holds across compute budgets; e.g., PAL and MAC stay at the top at lower budgets too (\appref{app:eval-optimizers-results}, \figref{fig:exp1-rank-budgets}).
Third, repeating the benchmark beyond LLMs, on corpus poisoning against four embedding models, also mostly agrees with the ranking and top optimizers in particular (Spearman's $\rho{=}0.94$; \appref{app:eval-optimizers-poisoning}, \figref{fig:exp1-emb-ranking}).

Overall, these results highlight the value of tracking discrete-optimizer progress, as enabled by \tropt{}, and motivate the use of underadopted methods (e.g., MAC and PAL)---which lead across compute budgets and in a second domain---in performance-critical settings, in particular security benchmarks and LLM red-teaming.

\subsection{Comparing Jailbreak Enhancements}
\label{sec:eval-jailbreak}

Beyond the optimizer itself, for LLM jailbreaks specifically, prior work proposes recipe-level enhancements---alternative losses, supplementary objectives, modified target responses, and specialized input templates---that aim to improve downstream attack success (\appref{app:eval-enh-setup}).
While each has been validated separately, the enhancements have never been compared head-to-head under controlled conditions.
We address this by fixing a generic recipe and applying each enhancement in isolation, measuring its contribution to jailbreak success.

\begin{wrapfigure}[15]{r}{0.51\linewidth}
    \centering
    \setlength{\abovecaptionskip}{2pt}
    \vspace{-2\baselineskip}
    \includegraphics[width=\linewidth]{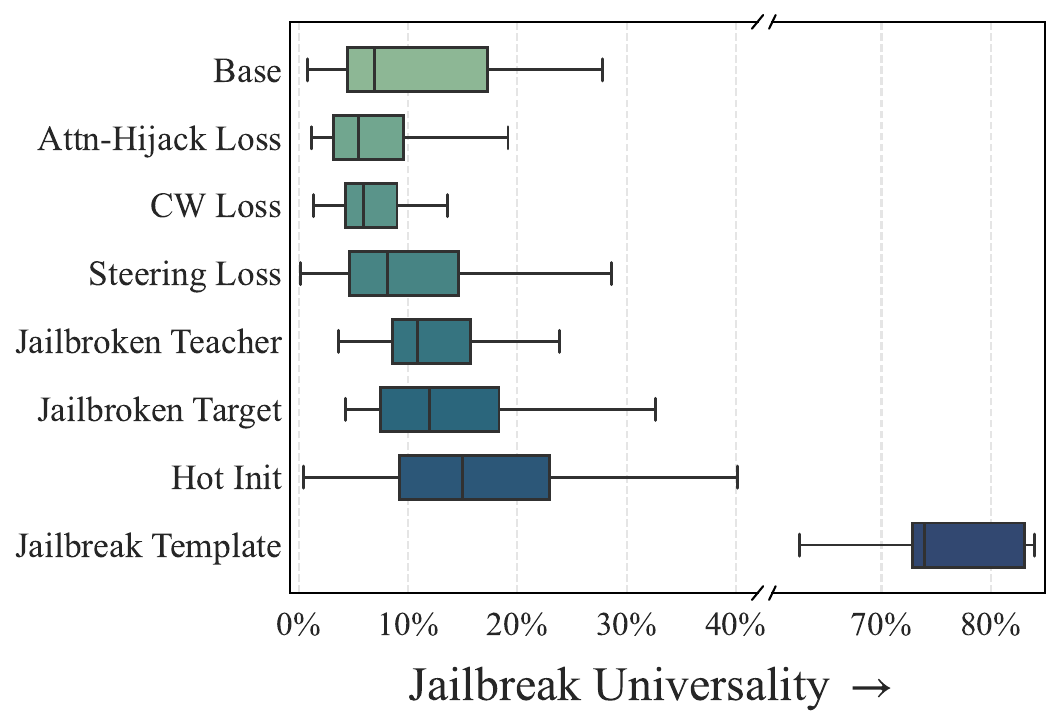}
    \caption{\textbf{Jailbreak Enhancement Comparison.}
    Comparing suffix universality per enhancement (rows), each applied to a generic \textit{Base} recipe.}
    \label{fig:exp2-succe}
    \vspace{1\baselineskip\relax}
\end{wrapfigure}

\parhead{Setup.}
We compare eight jailbreak enhancements from prior work~\citep[\textit{inter alia}]{thompson2024flrt,sitawarin2024palproxyguidedattack,huang2025irissuppressingrefusals}.
Following \secref{sec:eval-optimizers}, we target \gemmathree{} as a representative safety-aligned LLM \cite{gemma3}, optimizing against $15$ harmful instructions from ClearHarm~\citep{hollinsworth2025clearharm} across three seeds; our \textit{Base} recipe mirrors that of \secref{sec:eval-optimizers}, fixing the optimizer to MAC.
Each enhancement modifies exactly \textit{one} aspect of \textit{Base} (e.g., its loss or target string), isolating its effect.
We then score each crafted trigger by its \emph{universality}: the mean jailbreak success, per StrongReject-Finetuned model~\citep{souly2024strongreject}, over $100$ held-out ClearHarm instructions to which the trigger is appended. We defer further details to \appref{app:eval-enh-setup}.

\parhead{Results.}
\figref{fig:exp2-succe} plots the universality of each enhancement's crafted triggers, with \textit{Base} on top and the rest sorted by their median.
Modifying the loss---either by replacing it (e.g., CW loss) or by adding an auxiliary term (e.g., Attn-Hijack, Refusal-direction Steering)---brings no consistent gain over \textit{Base}: CW and Attn-Hijack land slightly below its median universality, and Steering only slightly above.
On the other hand, replacing the target string with the response of a jailbroken version of the target model---either from the response tokens (Jailbroken Target) or logits (Jailbroken Teacher)---raises the \textit{Base} median universality by over $1.5\times$ and shifts all its triggers' universality upward, confirming that the commonly used, canonical affirmative target is itself a bottleneck.
Warm-starting the trigger with a handcrafted jailbreak (Hot Init) raises the median further, to over $2\times$ that of \textit{Base}, but at the cost of inflated variance across runs.
Finally, the handcrafted jailbreak template of \citet{andriushchenko2024simpleadaptiveattacks}---which replaces the canonical \textit{instruction$\oplus$trigger} layout entirely---attains the highest universality, lifting all triggers above $60\%$. However, we find this success stems chiefly from the template itself rather than from the optimization
(\appref{app:eval-enh-eval}); moreover, its lengthy, explicit prompt layout may render the resulting attack conspicuous.
We defer additional results to \appref{app:eval-enh-eval}, including combinations
of enhancements and optimizing against \textit{multiple} harmful
instructions simultaneously to further encourage universality.

Overall, these results motivate red-teaming evaluations to adopt these more potent enhancements and to develop new ones along the same axes---both of which are now streamlined with \tropt{}.

\subsection{Cross-Domain Generalization of \tropt{}}\label{sec:eval-unifying}

To test \tropt{}'s ability to generalize optimizers beyond LLM jailbreaks, we compose three \emph{novel} recipes spanning different domains and model types, mixing and matching existing methods into unexplored combinations, and demonstrate their success in small-scale, proof-of-concept case studies.
\tropt{}'s modularity makes these adaptations---untried in the originating works---easy to realize, in turn surfacing new findings; e.g., we repurpose a black-box LLM-jailbreak optimizer into a successful corpus-poisoning attack on OpenAI's proprietary embedding model.
We defer setup details to \appref{app:eval-cross}.

\begin{figure}[t]
    \centering
    \begin{subfigure}[b]{0.44\linewidth}
        \centering
        \includegraphics[width=\linewidth]{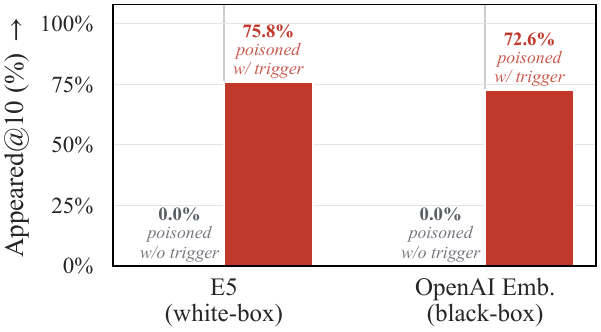}
        \caption{\textbf{Poisoning a Retrieval Corpus}}
        \label{fig:exp3-corpus-poisoning}
    \end{subfigure}
    \hfill
    \begin{subfigure}[b]{0.55\linewidth}
        \centering
        \includegraphics[width=\linewidth]{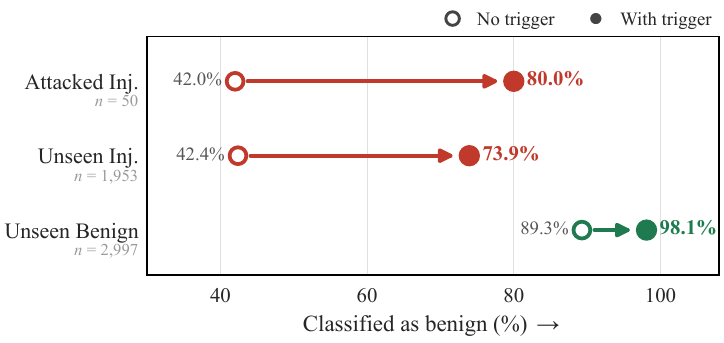}
        \caption{\textbf{Universal Trigger for Classifier Evasion}}
        \label{fig:exp3-adv-examples}
    \end{subfigure}
    \caption{\textbf{Cross-Domain Demonstrations of \tropt{}.}
    \textit{(a)} poisoning a corpus with ten passages---each appended with an optimized trigger---significantly lifts their appearance in the top-10 results for unseen queries.
    \textit{(b)} appending a universal optimized trigger to prompt injections successfully flips a detector's prediction on unseen samples.
    }
    \label{fig:exp3-cross-domain}
\end{figure}

\begin{itemize}
    \item \textbf{Corpus Poisoning Against Dense Retrievers.}
    Following the threat model of \citet{zhong2023poisoningretrieval}, we poison an 8M-passage retrieval corpus with 10 adversarial passages carrying malicious content,
    each optimized to rank in the top results for a target query set.
    Our recipe pairs the GASLITE optimizer with a cosine-similarity loss;
    for API-only models, we leverage \tropt{} to newly adapt the random search LLM-jailbreak optimizer by \citet{andriushchenko2024simpleadaptiveattacks}.
    Following \citet{BenTov2025GASLITE}, we target queries on Harry Potter, against the open-source E5 and the API-only OpenAI embeddings.\footnote{\href{https://hf.co/intfloat/e5-base-v2}{\texttt{intfloat/e5-base-v2}} and \href{https://platform.openai.com/docs/models/text-embedding-3-small}{\texttt{text-embedding-3-small}}}

    \figref{fig:exp3-corpus-poisoning} shows the optimized adversarial passages reach the top-10 for the majority of held-out queries on both models---marking, to our knowledge, the most successful embedding-output-only (API) corpus-poisoning attack tested against a proprietary embedding model.
    \item \textbf{A Universal Trigger for Evading a Prompt-Injection Classifier.}
    Building on \citet{wallace2019universaladversarialtriggers}, we
    craft a
    universal trigger that, appended to a prompt-injection message, flips a popular
    detector's\footnote{\url{https://huggingface.co/meta-llama/Llama-Prompt-Guard-2-86M}} prediction to benign.
    Our recipe pairs the GCG optimizer with a misclassification cross-entropy loss, optimizing the trigger across 50 prompt injections.

    \figref{fig:exp3-adv-examples} shows the optimized universal trigger generalizes to unseen prompt injections, evading the classifier in the vast majority of cases.
    \item \textbf{Prompt Recovery for Text-to-Image Models.}
    Following \citet{wen2023pez}, we recover a text prompt that, when fed to a text-to-image generator, regenerates a given image.
    Our recipe pairs the GCG optimizer---originally proposed for LLM jailbreaks---with a cosine-similarity loss against the image's CLIP embedding, using the frozen CLIP text encoder of the generator, Stable~Diffusion~2.1.

    \figref{fig:teaser}c shows the optimized, recovered prompt regenerates an image faithful to the original; additional examples in \tabref{tab:app-prompt-inv}.
 \end{itemize}

\section{Conclusion}
\label{sec:conclusion}

We introduce \tropt{}, an open-source framework for running and developing discrete text-trigger optimization strategies.
The key insight driving \tropt{} is that all discrete optimizers \textit{solve an identical algorithmic problem} across domains, yet engineering overhead has kept these techniques siloed---slowing adoption and obscuring comparison.
The stakes are highest in security red-teaming, where reliable robustness evaluation calls for stronger and rapidly adapted attacks (e.g., to stress-test a new defense)---both hindered by the current state of the field.
Addressing these hurdles, \tropt{} lets users, with minimal friction,
\textit{(i)} run dozens of existing optimizer applications out of the box (\secref{sec:tropt});
\textit{(ii)} compose new recipes, adapting any optimization scheme to any compatible model, application, and domain (\secref{sec:tropt-recipe});
and \textit{(iii)} customize recipes further by adding a new objective (\secref{sec:tropt-loss}) or optimizer (\secref{sec:tropt-optimizer}).
We demonstrate \tropt{}'s utility through a controlled optimizer comparison (\secref{sec:eval-optimizers}), an isolated evaluation of jailbreak enhancements (\secref{sec:eval-jailbreak}), and applications across different domains (\secref{sec:eval-unifying}).

We hope \tropt{} will help democratize discrete optimizers across existing and new applications,
enable tracking their progress through standardized benchmarks,
accelerate the development of adaptive attacks and new optimization strategies,
and ultimately advance defensive research in NLP models.
We intend to keep extending \tropt{} as the field evolves.

\parhead{Research Directions.}
\tropt{} streamlines existing lines of research and opens new ones.
First, the pre-configured optimization recipes directly support downstream studies---safety auditing, robustness analysis, defenses, interpretability---across LLMs, embedding models, and classifiers (\secref{sec:tropt-recipe}, \secref{sec:eval-unifying}).
Second, \tropt{}'s adaptability supports reliable red-teaming by lowering the barrier to forming stronger adversaries---e.g., via adaptive objectives or potent optimizers borrowed across domains.
Third, the empirical behavior of discrete optimizers (e.g., which fit which contexts) is underexplored; \tropt{}'s modular components lay the groundwork for standardized benchmarks across optimizers or other underexplored axes (e.g., \secrefs{sec:eval-optimizers}{sec:eval-jailbreak}).
Finally, \tropt{}'s infrastructure for optimizer development could serve as an efficient agentic harness for \emph{automated} optimizer discovery~\cite{panfilov2026claudini}.

\parhead{Limitations.}
\tropt{} targets \textit{discrete search} optimizers for neural text models, which directly search the input space via various strategies and heuristics.
Other trigger-optimization approaches---e.g., ad-hoc RL-trained LLMs or LLM-based agentic systems---currently fall outside its scope (see \secref{sec:scope}).
Our optimizer comparison (\secref{sec:eval-optimizers}) centers on LLM jailbreaks, with one extension to corpus poisoning against embedding models (\appref{app:eval-optimizers-poisoning}); evaluating optimizer potency more extensively---across further domains, with optimizer parameters tuned per setting and model, and under other comparison measures---remains to be explored, which \tropt{} readily enables. More broadly, we take a first step toward standardized benchmarking of discrete optimization methods, and defer more comprehensive comparisons to future work.

\ifisanon\else
\section*{Acknowledgments}

{\interlinepenalty=10000
This work has been supported in part
by a grant from the Blavatnik Interdisciplinary Cyber Research Center (ICRC);
by grant No.\ 2023641 from the United States-Israel Binational Science Foundation (BSF);
by an Intel Rising Star Faculty Award;
by Len Blavatnik and the Blavatnik Family Foundation;
by a Maof prize for outstanding young scientists;
by the Ministry of Innovation, Science \& Technology, Israel (grant number 0603870071);
by a Shashua scholarship for Ph.D.\ students; and
by a grant from the Tel Aviv University Center for AI and Data Science (TAD).
We thank Abdullah Garra for his assistance with experiments.
\par}
\fi

\bibliographystyle{plainnat}
\bibliography{bib}

\clearpage

\appendix

\section*{Ethical Considerations}

Our work introduces \tropt{}, an open-source framework that consolidates and democratizes discrete text-trigger optimization methods, many of which have been used by prior work to attack widely deployed text neural models---including LLMs, dense retrievers, and text classifiers.
While we aim to advance the security and analysis of such models, we recognize the potential for misuse; consolidating these methods lowers the barrier both to mounting such attacks and to developing new ones.
Prior to its release, we therefore disclosed \tropt{} and the attacks it enables to the affected model providers through responsible-disclosure channels.
The repository additionally includes responsible-use documentation, stating that \tropt{} is intended for defensive research and instructing users not to apply its techniques to deployed systems without authorization.
We have carefully considered the public release of \tropt{}'s codebase and believe the benefits outweigh the risks for the following reasons.

First, \tropt{} collects and re-implements already published methods, which a motivated attacker could readily reconstruct and combine---further boosted today by LLM-based coding tools.
An attacker, moreover, is seldom deterred by engineering overhead, and a loose reconstruction typically suffices; defensive research, in contrast, rests on reproducing prior methods faithfully and building on them.
The marginal uplift \tropt{} offers an attacker is therefore limited, while the uplift it offers defenders and researchers is substantial.
Similarly to our aims of advancing security research, prior work has organized and democratized attacks in open-source frameworks, including CleverHans~\citep{cleverhans}, ART~\citep{art}, Foolbox~\citep{foolbox}, and TextAttack~\citep{textattack}.

Second, \tropt{} offers a valuable tool for researchers and practitioners to reliably assess model robustness and develop defenses.
As repeatedly shown in machine-learning security literature~\citep{carlini2019evaluatingadversarialrobustness,nasr2025attackermovessecond}, defenses evaluated against weak or non-adaptive attacks risk a false sense of security;
democratizing access to potent, up-to-date optimizers---and streamlining their adaptation---thus drives more reliable security evaluation, and helps defenders surface and responsibly disclose vulnerabilities.

Finally, beyond red-teaming, \tropt{} supports a broad range of benign and analytical uses, including
toxicity auditing of LLMs~\citep{jones2023arcaauditingdiscrete},
studying memorization~\citep{schwarzschild2024rethinkingmemorization},
probing model internals~\citep{nikolaou2025injective},
and prompt recovery for text-to-image models~\citep{wen2023pez}.
Discrete optimizers also underpin a growing body of \emph{defensive} applications, such as adversarial training on optimized triggers~\citep{mazeika2024harmbench} and backdoor detection and extraction~\citep{shen2025bait}.

\section{Related Work}\label{app:related-work}

\parhead{Applications of \emph{Discrete Search} Optimizers.}
Following the rise of powerful neural text models,
discrete search text optimizers have gained reach across a wide range of research directions.
First, most prominently, they have exposed new \emph{inference-time attack vectors}:
adversarial examples against text classifiers~\cite{ebrahimi2018hotflip,wallace2019universaladversarialtriggers,guo2021gbda,davies2026bpj},
LLM jailbreaks~\citep[\textit{inter alia}]{zou2023universaltransferableadversarial,liu2023autodanjailbreak} and adaptive variants of them \cite{andriushchenko2024simpleadaptiveattacks,bailey2024obfuscatedactivations}, along with other LLM attack objectives~\cite{geiping2024coercing},
and corpus poisoning and embedding inversion against dense retrievers~\cite{zhong2023poisoningretrieval,zou2024poisonedrag,BenTov2025GASLITE,zhang2025zsinvert}.
Second, their automated and scalable nature has made them a standard tool for critical \emph{security evaluations}: jailbreak benchmarks~\cite{mazeika2024harmbench,chao2024jailbreakbench,nasr2025attackermovessecond},
unlearning evaluations~\cite{lucki2024adversarialunlearning},
and prompt-injection defenses~\cite{chen2025metasecalign}, all rely on discrete optimizers to surface failure modes.
Third, discrete optimizers are also central to the \emph{defender's} toolkit: adversarial training on optimized jailbreak triggers~\cite{mazeika2024harmbench},
backdoor detection and extraction~\cite{shen2025bait,rando2024competition}, safety auditing of LLMs~\cite{jones2023arcaauditingdiscrete}, and memorization auditing~\cite{schwarzschild2024rethinkingmemorization}.
Finally, they underpin a growing body of \emph{analysis and downstream applications}:
studies of robustness scaling and jailbreak side effects~\cite{howe2025scalingrobustness,nikolic2025jailbreaktax},
mechanistic interpretability of jailbreaks and refusal~\cite{arditi2024refusalmediatedsingle,ball2024understanding,bentov2025universaljailbreaksuffixesstrong},
probing LLM internals via hidden-state inversion~\cite{nikolaou2025injective},
auditing text-to-image models via prompt recovery~\cite{wen2023pez,chin2023prompting4debugging,williams2024promptrecovery},
and prompt tuning for downstream tasks~\cite{shin2020autoprompt}.

\parhead{Frameworks for Adversarial ML and Red-Teaming.}
\tropt{} builds on a lineage of open-source frameworks that have similarly consolidated adversarial ML research.
Vision adversarial-example libraries~\cite{cleverhans,art,foolbox} demonstrated how a common interface can organize a fast-growing family of attacks and support them as a standard evaluation tool; \tropt{} brings this approach to discrete text optimizers.
Textual adversarial-example frameworks~\cite{textattack,zeng2021openattack} applied this approach to NLP attacks; TextAttack, for instance, decomposes each attack into a goal function, transformations, constraints, and a search method, a modular design \tropt{} shares.
However, these frameworks predate modern trigger optimizers such as GCG and center on transformation-based methods against text classifiers, whereas \tropt{} hosts these optimizers and their successors across LLMs, embedding models, and classifiers.
Red-teaming benchmarks~\cite{mazeika2024harmbench,chao2024jailbreakbench} standardize how jailbreak attacks and defenses are \emph{evaluated}---fixing behaviors, judges, and threat models---and thus consume attacks as \textit{complete} methods; \tropt{} supports the lower-level infrastructure building these attacks,
and its recipes can be contributed back to these suites, thus accelerating their development.
Jailbreak toolkits consolidate LLM jailbreak attacks under a shared framework. EasyJailbreak~\cite{zhou2024easyjailbreak}, in particular, hosts 12 attacks built from prompt-level transformations (e.g., rewriting prompts) and scored on the model's textual responses; \tropt{} supports such jailbreak recipes and goes beyond them: it supports various types of objectives, on different models, using different techniques.

\section{\tropt{}: Additional Details}
\label{app:tropt}

As a concrete demonstration of the component design outlined in \secref{sec:tropt}, \coderef{code:custom-loss} and \coderef{code:custom-optimizer} show minimal custom loss and optimizer implementations, respectively.
We refer the reader to the quickstart notebook in \tropt{}'s codebase to instantly experiment with these custom components: \ifisanon\url{https://anonymous.4open.science/r/TROPT/quickstart.ipynb}\else\url{https://github.com/matanbt/TROPT/blob/main/quickstart.ipynb}\fi.

\begin{center}
\begin{minipage}{0.98\linewidth}
  \begin{lstlisting}[style=troptcode]
  from tropt.loss import BaseLoss
  from dataclasses import dataclass

  @dataclass
  class CustomSteeringLoss(BaseLoss):
      """Steers hidden states away from a refusal direction."""
      require_hidden_states = True  # requires model hidden-states computation
      refusal_dir = ...             # (d_model,)

      def __call__(
        self,
        full_hidden_states  # (bsz, n_layers, seq_len, d_model)
      ):
          h = full_hidden_states[:, -1, -1, :]  # (bsz, d_model); last layer and token
          return h @ self.refusal_dir           # (bsz,); minimize => steer away
  \end{lstlisting}
  \captionof{code}{\textbf{Implementing a custom loss} requires a short class, accepting standardized model outputs.
  In this example, the loss accepts the model hidden states and measures the alignment of the last layer and token with a specified direction.
  Crucially, \textit{any model} that provides hidden states will be \textit{compatible} with this loss out of the box, and this loss can be dropped into any recipe---including \coderef{code:gcg-composition}---requiring no other code changes.
  }
  \label{code:custom-loss}
\end{minipage}
\end{center}

\begin{code}[p]
  \centering
  \begin{minipage}{0.99\linewidth}
  \begin{lstlisting}[style=troptcode]
  from tropt.optimizer import BaseOptimizer, OptimizerResult
  from tropt.model import LossTokenAccessMixin
  import torch

  class CustomRandomOptimizer(BaseOptimizer):
    """Naive random search optimizer."""

    # requires target model to have token-level access to loss
    model_requirements = (LossTokenAccessMixin,)

    def __init__(
        self,
        model, loss, tracker=None, seed=None,
        # optimizer-specific parameters:
        num_steps=500, n_candidates=512,
    ):
        super().__init__(model, loss=loss, tracker=tracker, seed=seed)
        self.num_steps = num_steps
        self.n_candidates = n_candidates

    def optimize_trigger(self, templates, initial_trigger, targets):
        # register model inputs and targets
        self.model.set_inputs_from_tokens(templates, targets)

        # initialize trigger and loss
        best_trigger_ids = self.model.tokenizer.encode(
            initial_trigger, add_special_tokens=False
        )  # (trigger_len,)
        best_loss = float("inf")

        for step in self.track_steps(range(self.num_steps)):  # optionally caps FLOPs
            # sample fully random candidate triggers
            candidates = torch.randint(
                0, self.model.vocab_size,
                size=(self.n_candidates, len(best_trigger_ids)),
                device=self.model.device,
            )  # (n_candidates, trigger_len)

            # compute the loss of the inputs combined with the triggers
            # (handled internally in the model implementation)
            losses = self.model.compute_loss_from_tokens(
                candidates, self.loss_func
            )  # (n_candidates,)

            # update if improved
            best_cand = losses.argmin()
            if losses[best_cand] < best_loss:
                best_loss = losses[best_cand].item()
                best_trigger_ids = candidates[best_cand]

            # log step to the attached tracker (e.g., Wandb)
            self.log(loss=best_loss)

        return OptimizerResult(
            best_loss=best_loss,
            best_trigger_ids=best_trigger_ids,
            best_trigger_str=self.model.tokenizer.decode(best_trigger_ids),
        )
  \end{lstlisting}
  \end{minipage}
  \caption{\textbf{Implementing a custom optimizer} requires only filling in the core search scheme.
  Here,
  the optimizer declares its model requirement (loss computable from input tokens),
  then defines a self-contained search loop---a naive iterative random search.
  Within the loop, it tracks the best candidate and delegates loss computation to the \texttt{model} and \texttt{loss} components, while the framework handles step logging (e.g., streaming to a Wandb \texttt{tracker}).
  Crucially, this new optimizer composes with any compatible recipe (e.g., \coderef{code:gcg-composition}) with no other code changes.
  }
  \label{code:custom-optimizer}
  \end{code}

\section{\tropt Component Catalog}
\label{app:catalog}

We provide a comprehensive catalog of the components currently available in \tropt.
\tabref{tab:attack-zoo} lists selected pre-configured recipes available in the Recipe Hub;
\tabref{tab:optimizers} details the optimization algorithms;
and \tabref{tab:losses} describes the loss functions.

\begin{table}[p]
\centering
\caption{\textbf{Pre-Configured \tropt Recipes}, each composing an optimizer and a loss into a runnable attack; beneath each attack name is the recipe's exact Recipe-Hub entry point, invocable as \recipefn{tropt.recipe\_hub.\textit{name}}.
\textbf{Type} follows \tropt's naming convention---\textbf{R}: a \emph{reproduction} of the reference's algorithm and hyperparameters; \textbf{A}: a \tropt \emph{adaptation}, recomposing components in a way the reference did not.
Access requirements are properties of the optimizer and loss (\tabrefs{tab:optimizers}{tab:losses}).}
\label{tab:attack-zoo}
\scriptsize
\setlength{\tabcolsep}{2.5pt}
\renewcommand{\arraystretch}{0.75}
\setlength{\aboverulesep}{0.25ex}
\setlength{\belowrulesep}{0.4ex}
\begin{tabularx}{\textwidth}{@{}>{\raggedright\arraybackslash}p{3.05cm} >{\raggedright\arraybackslash}p{2.76cm} >{\raggedright\arraybackslash}p{2.90cm} >{\raggedright\arraybackslash}X c@{}}
\toprule
\textbf{Attack} & \textbf{Optimizer} & \textbf{Loss} & \textbf{Reference} & \textbf{Type} \\
\midrule
\multicolumn{5}{l}{\textit{Targeting LLM for Jailbreak; Gradient-Based}} \\
\midrule
\attackrecipe{HotFlip}{hotflip\_\_\rbk ebrahimi2018}
  & \texttt{HotFlipOptimizer}      & \texttt{PrefillCELoss}                                    & \citet{ebrahimi2018hotflip}                             & R \\ \addlinespace[1pt]
\attackrecipe{AutoPrompt}{autoprompt\_\_\rbk shin2020}
  & \texttt{AutoPromptOptimizer}   & \texttt{PrefillCELoss}                                    & \citet{shin2020autoprompt}                              & R \\ \addlinespace[1pt]
\attackrecipe{GBDA}{gbda\_\_\rbk guo2021}
  & \texttt{GBDAOptimizer}         & \texttt{PrefillCELoss}                                    & \citet{guo2021gbda}                                     & R \\ \addlinespace[1pt]
\attackrecipe{ARCA}{arca\_\_\rbk jones2023}
  & \texttt{ARCAOptimizer}         & \texttt{PrefillCELoss}                                    & \citet{jones2023arcaauditingdiscrete}                   & R \\ \addlinespace[1pt]
\attackrecipe{GCG}{gcg\_\_\rbk zou2023}
  & \texttt{GCGOptimizer}          & \texttt{PrefillCELoss}                                    & \citet{zou2023universaltransferableadversarial}          & R \\ \addlinespace[1pt]
\attackrecipe{PEZ}{pez\_\_\rbk wen2023}
  & \texttt{PEZOptimizer}          & \texttt{PrefillCELoss}                                    & \citet{wen2023pez}                                      & R \\ \addlinespace[1pt]
\attackrecipe{MAC}{mac\_\_\rbk zhang2024}
  & \texttt{GCGPlusOptimizer}      & \texttt{PrefillCELoss}                                    & \citet{zhang2024mac}                                    & R \\ \addlinespace[1pt]
\attackrecipe{GCG-Hij}{gcg\_\rbk hij\_\_\rbk bentov2025}
  & \texttt{GCGOptimizer}          & \texttt{PrefillCELoss} + \texttt{AttentionEnhLoss}        & \citet{bentov2025universaljailbreaksuffixesstrong}      & R \\ \addlinespace[1pt]
\attackrecipe{IRIS}{iris\_\_\rbk huang2025}
  & \texttt{GCGOptimizer}          & \texttt{PrefillCELoss} + \texttt{SteeringActivationLoss}  & \citet{huang2025irissuppressingrefusals}                & R \\
\midrule
\multicolumn{5}{l}{\textit{Targeting LLM for Jailbreak; Query-Based}} \\
\midrule
\attackrecipe{PAL}{pal\_\_\rbk sitawarin2024}
  & \texttt{PALOptimizer}          & \texttt{PrefillCELoss}                                    & \citet{sitawarin2024palproxyguidedattack}               & R \\ \addlinespace[1pt]
\attackrecipe{RAL}{ral\_\_\rbk sitawarin2024}
  & \texttt{PALOptimizer}          & \texttt{PrefillCELoss}                                    & \citet{sitawarin2024palproxyguidedattack}               & R \\ \addlinespace[1pt]
\attackrecipe{QCG}{qcg\_\_\rbk hayase2024}
  & \texttt{QCGOptimizer}          & \texttt{PrefillCELoss}                                    & \citet{hayase2024gcqquerybased}                         & R \\ \addlinespace[1pt]
\attackrecipe{Prompt Random Search}{prs\_\_\rbk andriushchenko2024}
  & \texttt{RandomSearchOptimizer} & \texttt{FirstTokenNLLLoss}                                & \citet{andriushchenko2024simpleadaptiveattacks}         & R \\ \addlinespace[1pt]
\attackrecipe{BEAST}{beast\_\_\rbk sadasivan2024}
  & \texttt{BeamSearchOptimizer}   & \texttt{PrefillCELoss}                                    & \citet{sadasivan2024beast}                              & R \\ \addlinespace[1pt]
\attackrecipe{AdvDecoding}{advdecoding\_\rbk jailbreak\_\_\rbk zhang2024}
  & \texttt{BeamSearchOptimizer}   & \texttt{PrefillCELoss} + \texttt{InputFluencyLoss}        & \citet{zhang2025advdecoding}                            & R \\
\midrule
\multicolumn{5}{l}{\textit{Targeting Embedding Models for Corpus Poisoning}} \\
\midrule
\attackrecipe{GASLITE}{gaslite\_\_\rbk bentov2024}
  & \texttt{GASLITEOptimizer}      & \texttt{SimilarityLoss}                                   & \citet{BenTov2025GASLITE}                               & R \\ \addlinespace[1pt]
\attackrecipe{Random Search (Ret.)}{rs\_\rbk emb}
  & \texttt{RandomSearchOptimizer} & \texttt{SimilarityLoss}                                   & \citet{andriushchenko2024simpleadaptiveattacks}         & A \\ \addlinespace[1pt]
\attackrecipe{AdvDecoding (Ret.)}{advdecoding\_\rbk retrieval\_\_\rbk zhang2024}
  & \texttt{BeamSearchOptimizer}   & \texttt{SimilarityLoss} + \texttt{InputFluencyLoss}       & \citet{zhang2025advdecoding}                            & R \\
\midrule
\multicolumn{5}{l}{\textit{Text-to-Image Model Auditing}} \\
\midrule
\attackrecipe{Prompt Recovery}{prompt\_\rbk recovery\_\_\rbk wen2023}
  & \texttt{PEZOptimizer}          & \texttt{SimilarityLoss}                                   & \citet{wen2023pez,williams2024promptrecovery}           & R \\
\midrule
\multicolumn{5}{l}{\textit{LLM Safety Auditing}} \\
\midrule
\attackrecipe{Toxic Comments}{arca\_\rbk toxic\_\rbk reverse}
  & \texttt{GCGPlusOptimizer}      & \texttt{PrefillCELoss}                                    & \citet{jones2023arcaauditingdiscrete}                   & A \\
\midrule
\multicolumn{5}{l}{\textit{Targeting Classifier for Adversarial Examples}} \\
\midrule
\attackrecipe{Classifier GCG}{classifier\_\rbk gcg}
  & \texttt{GCGOptimizer}          & \texttt{MisclassCELoss}                                   & ---                                                     & A \\ \addlinespace[1pt]
\attackrecipetwo{UAT}{uat\_\rbk classifier}{uat\_\rbk prompt\_\rbk injection}
  & \texttt{GCGPlusOptimizer}      & \texttt{MisclassCELoss}                                   & \citet{wallace2019universaladversarialtriggers}         & A \\
\bottomrule
\end{tabularx}
\end{table}

\begin{table}[t]
\centering
\caption{\textbf{Optimization Algorithms in \tropt{}, and What Each Requires of the Model.}
\textbf{Optimizer} is the \tropt class, grouped by search strategy; \textbf{Instantiates} lists the published attacks it implements.
\textbf{Requires} states any requirement on the victim that goes \emph{beyond} the basic API every model provides: textual input, the model's default output (e.g., a text response for an LLM, or the final vector for an embedding model).
Requirements are additive with the loss's (\tabref{tab:losses}): a gradient-free optimizer under a prefill loss still needs prefilled-response logits.}
\label{tab:optimizers}
\footnotesize
\setlength{\tabcolsep}{4pt}
\renewcommand{\arraystretch}{0.95}
\newcommand{\nosup}{\textcolor{gray}{---}}
\begin{tabularx}{\textwidth}{@{}>{\raggedright\arraybackslash}p{3.5cm} >{\raggedright\arraybackslash}X >{\raggedright\arraybackslash}p{4.3cm}@{}}
\toprule
\textbf{Optimizer} & \textbf{Instantiates} & \textbf{Requires} \\
\midrule
\multicolumn{3}{@{}l}{\textit{Gradient-based discrete} --- back-propagates through the loss, which must be differentiable} \\
\midrule
\texttt{GCGOptimizer}          & \mbox{GCG~\citep{zou2023universaltransferableadversarial}}   & Token inputs and gradients \\
\texttt{GCGPlusOptimizer}      & \mbox{GCG+~\citep{sitawarin2024palproxyguidedattack}} \newline
                                 \mbox{GCG+~\citep{hayase2024gcqquerybased}} \newline
                                 \mbox{MAC~\citep{zhang2024mac}} \newline
                                 \mbox{UAT~\cite{wallace2019universaladversarialtriggers}}    & Token inputs and gradients \\
\texttt{AutoPromptOptimizer}   & \mbox{AutoPrompt~\citep{shin2020autoprompt}}                 & Token inputs and gradients \\
\texttt{HotFlipOptimizer}      & \mbox{HotFlip~\citep{ebrahimi2018hotflip}}                   & Token inputs and gradients \\
\texttt{ARCAOptimizer}         & \mbox{ARCA~\citep{jones2023arcaauditingdiscrete}}            & Token inputs and gradients \\
\texttt{GASLITEOptimizer}      & \mbox{GASLITE~\citep{BenTov2025GASLITE}}                     & Token inputs and gradients \\
\texttt{PALOptimizer}          & \mbox{PAL~\citep{sitawarin2024palproxyguidedattack}}         & Token inputs and gradients \\
\quad w/ proxy grad.\ model    &                                                              & Token inputs \\
\midrule
\multicolumn{3}{@{}l}{\textit{Continuous relaxation} --- back-propagates through the loss, which must be differentiable} \\
\midrule
\texttt{GBDAOptimizer}         & \mbox{GBDA~\citep{guo2021gbda}}                              & Embedding inputs and gradients \\
\texttt{PEZOptimizer}          & \mbox{PEZ~\citep{wen2023pez}}                                & Embedding inputs and gradients \\
\midrule
\multicolumn{3}{@{}l}{\textit{Zeroth order} --- only evaluates the loss, so any loss applies} \\
\midrule
\texttt{RandomSearchOptimizer} & \mbox{PRS~\citep{andriushchenko2024simpleadaptiveattacks}}   & \nosup \\
\texttt{BeamSearchOptimizer}   & \mbox{BEAST~\citep{sadasivan2024beast}}                      & Token inputs and full logits \\
\quad w/ auxiliary LLM         & \mbox{AdvDecoding~\citep{zhang2025advdecoding}}              & \nosup \\
\texttt{PALOptimizer}, as RAL  & \mbox{RAL~\citep{sitawarin2024palproxyguidedattack}}         & Token inputs \\
\texttt{QCGOptimizer}          & \mbox{QCG~\citep{hayase2024gcqquerybased}}                   & \nosup \\
\bottomrule
\end{tabularx}
\end{table}

\begin{table}[t]
\centering
\caption{\textbf{Loss Functions in \tropt{}, and the Signal Each Consumes.}
\textbf{Operates On} is the model signal the loss reads and requires access to, \emph{beyond} the basic text-input and default output API (e.g., textual response in LLMs).
A loss's requirement is additive with the optimizer's.
\textbf{Objective} states what the loss optimizes.}
\label{tab:losses}
\footnotesize
\setlength{\tabcolsep}{3pt}
\renewcommand{\arraystretch}{0.9}
\newcommand{\nosup}{\textcolor{gray}{---}}
\begin{tabularx}{\textwidth}{@{}l >{\raggedright\arraybackslash}p{3.0cm} >{\raggedright\arraybackslash}X@{}}
\toprule
\textbf{Loss} & \textbf{Operates On} & \textbf{Objective} \\
\midrule
\multicolumn{3}{@{}l}{\textit{Logit-Based (Prefill)}} \\
\midrule
\texttt{PrefillCELoss}
  & Full resp.\ logits
  & Maximize likelihood of a target response~\citep{zou2023universaltransferableadversarial} \\
\texttt{PrefillCWLoss}
  & Full resp.\ logits
  & Push target logits above all others by a margin~\citep{sitawarin2024palproxyguidedattack,carlini2017cw} \\
\texttt{PrefillDistillationLoss}
  & Full resp.\ logits
  & KL divergence with target (teacher) logits~\cite{thompson2024flrt} \\
\midrule
\multicolumn{3}{@{}l}{\textit{Embedding-Based}} \\
\midrule
\texttt{SimilarityLoss}
  & \nosup
  & Maximize cosine similarity with a target vector \\
\midrule
\multicolumn{3}{@{}l}{\textit{Model Internal-Based}} \\
\midrule
\texttt{AttentionEnhLoss}
  & Attn.\ scores \newline \emph{(input tokens only)}
  & Maximize attention along a given flow~\citep{wang2024attngcg,bentov2025universaljailbreaksuffixesstrong} \\
\texttt{SteeringActivationLoss}
  & Internal activations \newline \emph{(input tokens only)}
  & Steer activations toward/away from a target direction~\citep{huang2025irissuppressingrefusals} \\
\midrule
\multicolumn{3}{@{}l}{\textit{Classification-Based}} \\
\midrule
\texttt{MisclassCELoss}
  & Class logits
  & Minimize/maximize target class probability \\
\midrule
\multicolumn{3}{@{}l}{\textit{Text-Based (Non-Differentiable)}} \\
\midrule
\texttt{FirstTokenNLLLoss}
  & 1st-tok.\ logprobs
  & NLL of a target token (e.g., ``Sure'') in the first generated token~\citep{andriushchenko2024simpleadaptiveattacks} \\
\texttt{ResponseHarmfulnessLoss}
  & \nosup
  & LLM-judge score for response harmfulness \\
\midrule
\multicolumn{3}{@{}l}{\textit{Meta}} \\
\midrule
\texttt{CombinedLoss}
  & (per component)
  & Weighted sum of multiple losses; enables multi-objective optimization \\
\bottomrule
\end{tabularx}
\end{table}

\section{Reproducing GCG with \tropt{}}
\label{app:reproducing-gcg}

To validate that our framework implementation is faithful to existing, commonly used implementations, we test it head-to-head against nanoGCG, a popular standalone implementation of GCG.\footnote{\url{https://github.com/GraySwanAI/nanoGCG}}

\begin{figure}[t]
    \centering

    \begin{subfigure}[b]{\linewidth}
        \centering
        \includegraphics[width=0.9\linewidth]{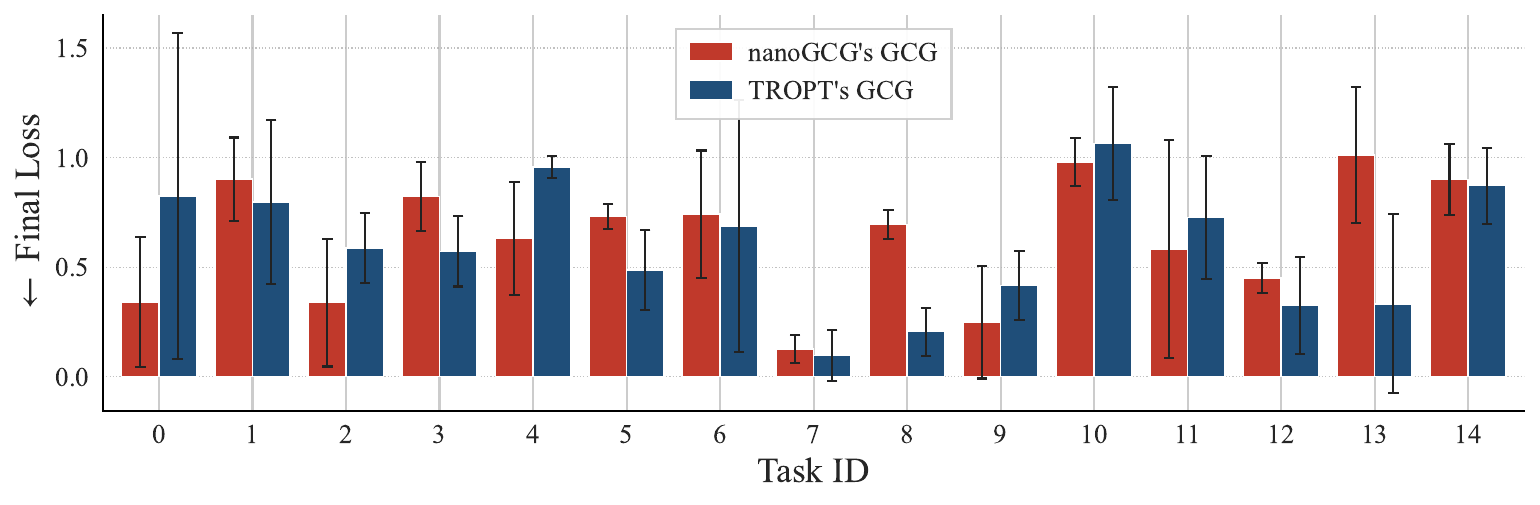}
        \caption{Final Loss Per Instruction (avg. on three seeds)}
        \label{fig:reprod-final}
    \end{subfigure}

    \vspace{0.5em}
    \begin{subfigure}[b]{\linewidth}
        \centering
        \includegraphics[width=\linewidth]{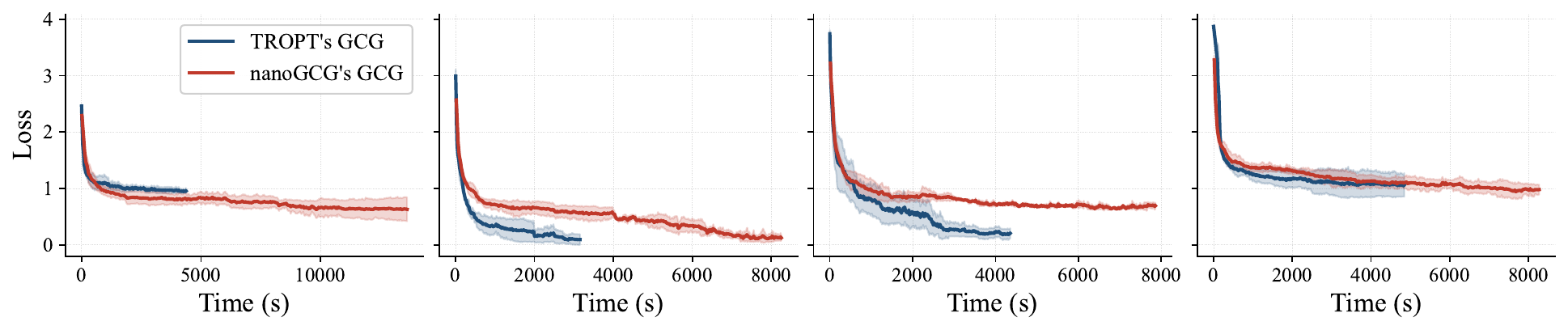}
        \caption{Loss vs. Runtime (seconds)}
        \label{fig:reprod-dynamics}
    \end{subfigure}

    \caption{\tropt's GCG vs.\ nanoGCG on \gemmathree{}, under matched hyperparameters and the same number of optimization steps.
    \tropt (a) reaches a comparable loss on average across instructions,
    and (b) finishes the same $500$-step GCG runs $2.5\times$ faster on average.}
    \label{fig:reprod-gcg}
\end{figure}

\parhead{Setup.}
We mirror the setup of \secref{sec:eval-optimizers}, replacing the optimizer being benchmarked with each of the two GCG implementations.
Specifically, we target \gemmathree on $15$ harmful instructions from ClearHarm,
each repeated over three random seeds, yielding $45$ paired (instruction, seed) tasks per implementation.
Both implementations are configured with identical, original GCG hyperparameters: $500$ optimization steps, $512$ candidates per step, top-$256$ token sampling, a single token replacement per step, a randomly initialized $20$-token suffix, the same per-seed initialization, and the same target prefill (\texttt{PrefillCE} loss).
We run the experiments on a single GPU of \textit{NVIDIA RTX A6000} with 48GB VRAM.

\parhead{Results.}
\figref{fig:reprod-final} shows the average loss at the granularity of instructions,
while \figref{fig:reprod-dynamics} shows the loss dynamics as a function of the runtime for four randomly sampled instructions (each averaged across the three seeds).
Across the $45$ paired tasks, the two implementations reach essentially
the same final loss on average (\tropt: $0.597 \pm 0.384$; nanoGCG: $0.633 \pm 0.335$),
with \tropt{}'s implementation surpassing nanoGCG's in $25/45$ of the tasks.
We spot a clear difference in efficiency: although both are set to run the same number of steps ($500$), \tropt{}'s implementation takes
$\sim$$2.5\times$ less runtime than nanoGCG, with $\sim$60 vs.\ $\sim$149 minutes on average.
We note that the algorithm itself is unchanged across implementations, and both fit the largest batch per forward pass.
We attribute the speedup to an accumulation of engineering optimizations we make in \tropt's implementation.\footnote{For instance, nanoGCG calls \texttt{torch.cuda.empty\_cache()} repeatedly during dynamic batching, incurring significant runtime overhead; \tropt{} batches dynamically too but avoids this overhead.}
Overall, this confirms that \tropt's GCG faithfully reproduces nanoGCG's optimization quality.

\section{Benchmarking Optimization Strategies: Additional Details and Results}
\label{app:eval-optimizers}

This section extends \secref{sec:eval-optimizers},
providing
additional details on the experimental setup, and
supplementary analyses of the results.

\subsection{Detailed Setup}\label{setup-optimizers}

\parhead{Optimizer Selection.}
Our selection of 14 optimizers spans diverse algorithmic families: gradient-based search (e.g., GCG, GASLITE, PAL) and its gradient-free variants (e.g., RAL, QCG); random-search heuristics (Random Search); continuous relaxation of the input embeddings (PEZ, GBDA); and beam search over LLM sampling (BEAST, AdvDecoding).
We additionally include optimizers from seminal work in the field (HotFlip, AutoPrompt), and favor widely used ones (e.g., those adopted by popular red-teaming benchmarks).
This selection is not an exhaustive coverage of the many existing optimizers, as we prioritize implementing and reproducing each one reliably over their number; \tropt{}'s extensibility (\secref{sec:tropt-optimizer}) makes adding further optimizers straightforward.

\parhead{Evaluated Optimizers.}
For the evaluation we set the following recipe across all optimizers: we use \texttt{PrefillCE} as a loss against a target response from the dataset; the trigger length is set to $T=20$; the trigger is randomly initialized; we disable non-ASCII and special tokens; below we list each optimizer instantiation, noting the algorithm-specific hyperparameters.
Each optimizer runs until the FLOP limit ($3\times 10^{17}$; adopting the counter by \citealt{boreiko2024realisticthreatmodel}) is exhausted, unless the optimizer forces an early stop by design; e.g., BEAST and AdvDecoding both sample from an LLM $T$ times, thus finish before exhausting this FLOP limit.
For each model, we perform $45$ runs per optimizer: $15$ harmful instructions $\times$ $3$ random seeds. The $15$ instructions are randomly sampled from ClearHarm \cite{hollinsworth2025clearharm}, but consistent across optimizers and models. Diverse affirmative target response strings were generated using Claude Opus 4.6.
We run the experiments on a single \textit{NVIDIA RTX A6000} with 48GB VRAM, with the only exception being runs with \gemmaFour{}, which run on a single \textit{NVIDIA H100} with 80GB VRAM.
For the measurement, for each model, we rank the optimizers on each run (i.e., a specific instruction and seed) according to the best loss they obtain throughout the optimization. Finally, we average the ranks of each optimizer across runs, yielding the \textit{Mean Rank} of optimizers per model.
\ifisanon\else We release the optimized triggers of all runs, along with their generated responses and full evaluation for future research.\footnote{\href{https://huggingface.co/datasets/MatanBT/tropt-jb-optbench}{\texttt{hf.co/datasets/MatanBT/tropt-jb-optbench}}}\fi

\begin{itemize}
    \item \textbf{HotFlip}~\citep{ebrahimi2018hotflip}:
    iteratively picks the best single-token flip according to the gradient, without loss evaluation.
    \item \textbf{AutoPrompt}~\citep{shin2020autoprompt}:
    gradient-based candidate sampling, followed by the candidates' loss evaluation;
    adapting for LLM jailbreak, we align parameters with GCG's (as done by \citealt{zou2023universaltransferableadversarial}): the candidate sample size is set to $512$, selected from the top-$256$ token ids per position.
    \item \textbf{GBDA}~\citep{guo2021gbda}:
    continuous relaxation with Gumbel-softmax sampling and gradient access;
    to match the original paper, we set
    $10$ gradient samples, learning rate $0.3$; the final trigger is chosen as the lowest-loss one among $100$ final Gumbel samples; no temperature annealing or LR decay.
    \item \textbf{ARCA}~\citep{jones2023arcaauditingdiscrete}:
    averaged-gradient based candidate sampling, followed by their loss evaluation; matching the original paper we take the gradient average over $32$ samples;
    similarly to AutoPrompt, we align ARCA with GCG's parameters: $512$ candidates, top-$256$ token sampling.
    \item \textbf{PEZ}~\citep{wen2023pez}:
    continuous-embedding optimization with discrete projection each step;
    matching the original paper we set learning rate $0.1$, weight decay $0.1$.
    \item \textbf{GCG}~\citep{zou2023universaltransferableadversarial}:
    gradient-guided candidate sampling;
    $512$ candidates over top-$256$ token sampling,
    with retokenization filtering.
    \item \textbf{MAC}~\citep{zhang2024mac}: Adds gradient momentum on top of GCG.
    Following the paper we use $\mu=0.6$ as the momentum coefficient, $256$ candidates, top-$256$ token sampling, with retokenization filtering.
    \item \textbf{GASLITE}~\citep{BenTov2025GASLITE}:
    Gradient-based multi-coordinate candidate sampling with gradient averaging.
    We set gradient averaging over $10$ samples, and based on the gradient we take $7$ flips per step, each flip evaluates the $256$ top candidates, with retokenization filtering.
    \item \textbf{PAL}~\citep{sitawarin2024palproxyguidedattack}:
    Adds several modifications to GCG, and enables a proxy-guided targeting of black-box models; since we target a white-box model, our proxy model is set to be the target model itself. Per the original paper, it uses $128$ candidates over top-$256$ token sampling, with retokenization filtering.
    \item \textbf{RAL}~\citep{sitawarin2024palproxyguidedattack}:
    PAL ablation that replaces the gradient with a random tensor for candidate selection, making it a black-box attack;
    uses $32$ random candidates per step, with retokenization filtering.
    \item \textbf{Random Search}~\citep{andriushchenko2024simpleadaptiveattacks}:
    A black-box, zeroth-order attack that operates through iterative block-random mutation;
    starting from block length $4$, decayed throughout the steps (following the paper's decay scheme), it randomly mutates a contiguous block at random positions, creating $128$ candidates per step; resets the whole trigger after $50$ steps of patience (no improvement in loss).
    \item \textbf{BEAST}~\citep{sadasivan2024beast}:
    A black-box attack that samples the trigger tokens from the target LLM's own next-token distribution using beam-search, while minimizing the target loss.
    Per the original paper we set beam size $15$, branching $15$, and sample over the full token distribution (up to token constraints); the method returns the trigger after autoregressively sampling $T$ tokens, thus finishes before our FLOP limit.
    \item \textbf{QCG}~\citep{hayase2024gcqquerybased}:
    A black-box, zeroth-order attack.
    Iteratively samples $1024$ candidates by making a single random token flip on each (reduced from $8192$ in the original paper to reduce compute), retokenizes them and evaluates them on the target model, while maintaining a buffer of $128$ best triggers.
    \item \textbf{AdvDecoding}~\citep{zhang2025advdecoding}:
    A black-box attack that samples the trigger tokens from an \textit{auxiliary} LLM's next-token distribution using beam-search, while minimizing the target loss.
    We use \texttt{google/gemma-2-2b-it} as the auxiliary LLM, with beam size $96$, branching $10$, and top-$k=10$ sampling. Since AdvDecoding is the only optimizer that relies on an auxiliary LLM, we count its compute toward the FLOP cap; in practice, AdvDecoding does not exhaust this cap, as it takes $T$ steps to finish.
\end{itemize}

\subsection{Additional Results}
\label{app:eval-optimizers-results}

\parhead{Mean Ranking Statistical Tests.}
To reflect the statistical significance of our benchmark comparison we run two complementary statistical tests \cite{demsar06a}, which originally motivated our rank-based analysis in \secref{sec:eval}.
First, we run a Friedman test across the $N=180$ tasks (provided by $4$ models $\times$ $15$ instructions $\times$ $3$ seeds) and $K=14$ optimizers, and it rejects the null hypothesis of equal optimizer performance ($\chi^2=1468,\ p<10^{-300}$).
Then, running a post-hoc Nemenyi test at $\alpha=0.05$ yields a critical difference of $\mathrm{CD}=1.48$ ranks. This means that, within this CD threshold, the leading PAL and MAC form a statistically indistinguishable group, of which only PAL significantly outperforms the canonical GCG baseline, which in turn is comparable to the gradient-free RAL; HotFlip, on the other end, is significantly worse than \emph{any} other optimizer.

\begin{figure*}[h!]
    \centering
    \includegraphics[width=0.95\linewidth]{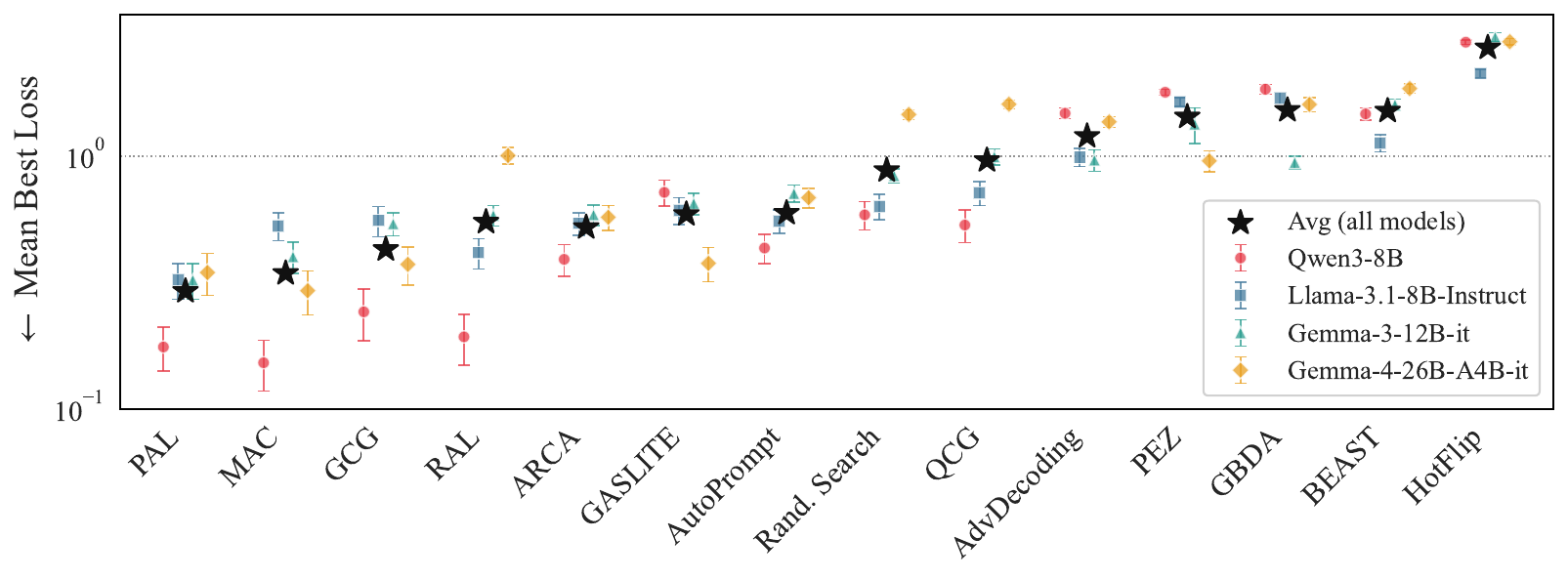}
    \caption{\textbf{Per-Model Mean Best Loss.} Per-model mean best loss for each optimizer (lower is better), sorted by the loss-based \emph{Mean Rank} used in the main evaluation (\figref{fig:exp1-ranking}); black $\bigstar$ marks the average across all models, and error bars the standard error of the mean.}
    \label{fig:exp1-loss-scatter}
\end{figure*}

\parhead{Per-Model Mean Best Loss.}
\figref{fig:exp1-loss-scatter} mirrors \figref{fig:exp1-ranking} but reports absolute loss values (log scale) rather than ranks, allowing comparison of both relative ordering and loss magnitude across models. Each loss value is calculated w.r.t. the optimized instruction and trigger. Error bars show the standard error of the mean across seeds and instructions.

\begin{figure}[h!]
    \centering
    \includegraphics[width=0.95\linewidth]{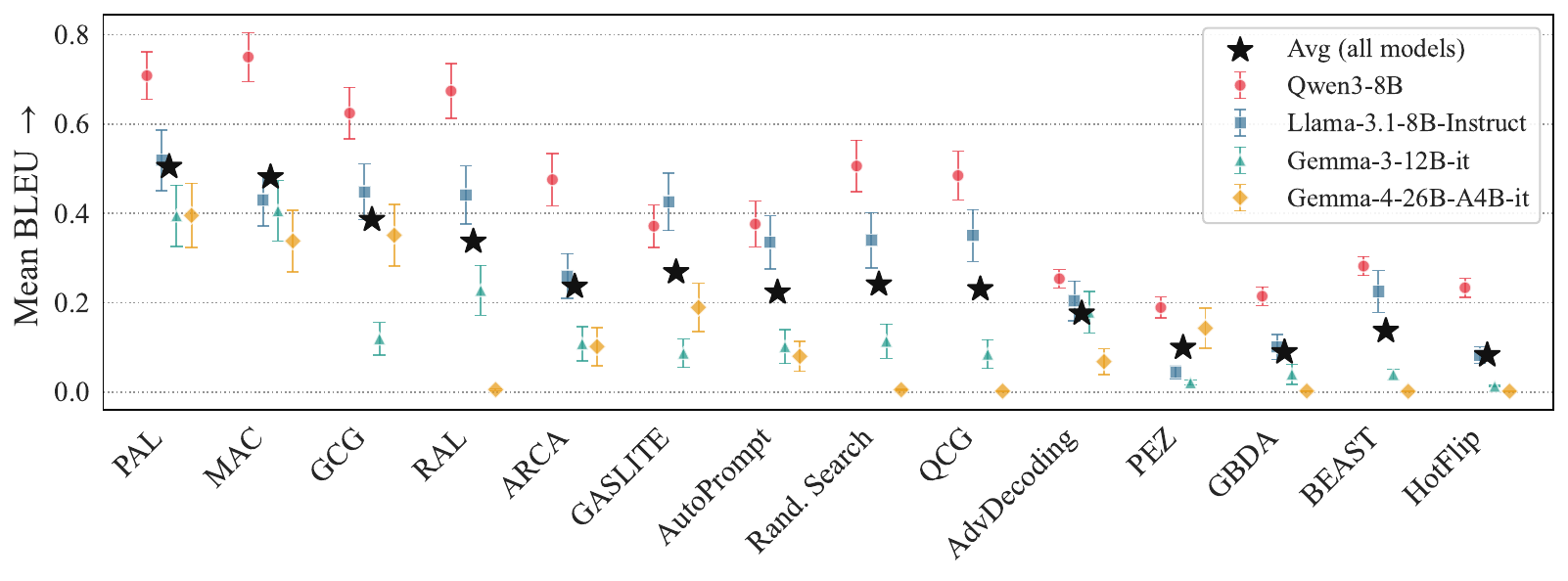}
    \caption{\textbf{Per-Model Mean BLEU Score.} Per-model mean BLEU between the target response string and the generated response with the optimized triggers for each optimizer (higher is better), sorted by the loss-based \emph{Mean Rank} (black $\bigstar$: average across all models; error bars: standard error of the mean). The average BLEU strongly correlates with the loss-based \emph{Mean Rank} in \figref{fig:exp1-ranking} (Spearman's $\rho=-0.96$).}
    \label{fig:exp1-bleu}
\end{figure}

\parhead{Per-Model Mean BLEU Score.}
\figref{fig:exp1-bleu} reports the average BLEU score of the responses to the optimized instruction and trigger per optimizer, with the standard error of the mean across seeds and instructions. BLEU~\cite{papineni2002bleu} measures the n-gram overlap between the target response and the response the model actually generates for the optimized prompt; unlike the loss, which scores the prefilled target, it thus verifies that the optimization indeed elicits the target in generation.

\begin{figure}[h!]
    \centering
    \includegraphics[width=0.95\linewidth]{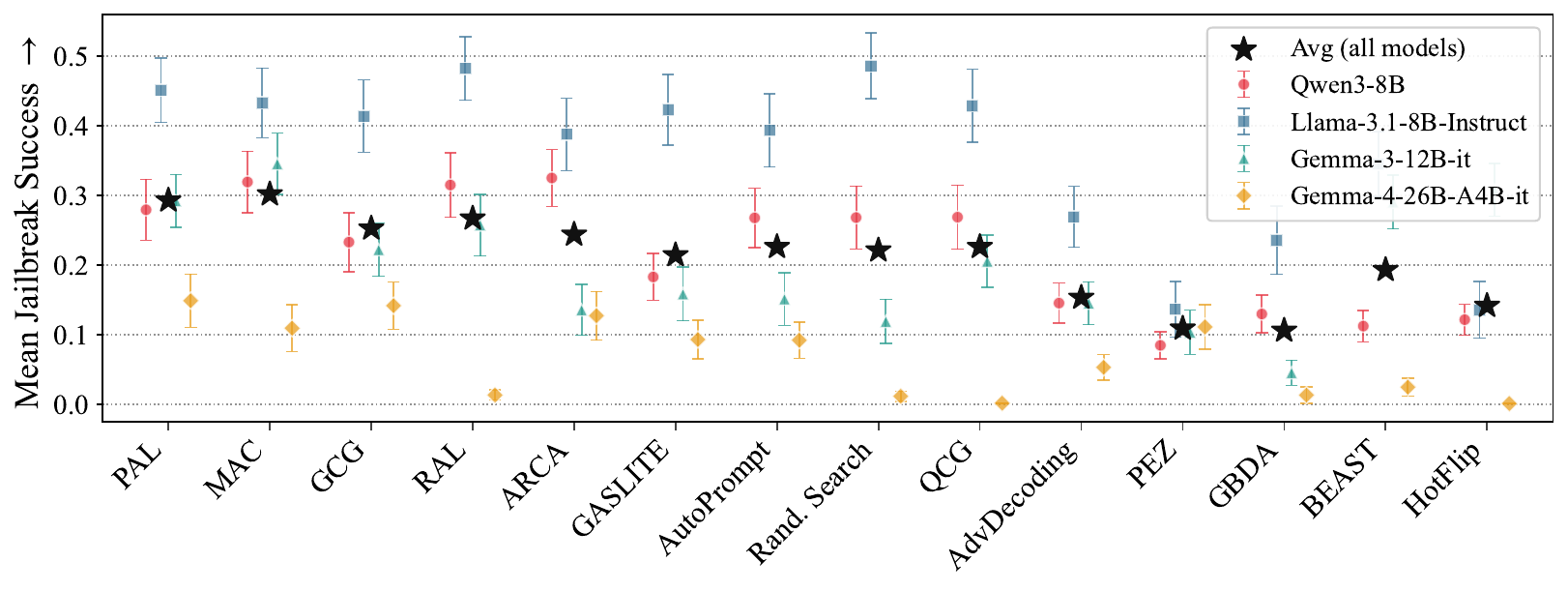}
    \caption{\textbf{Per-Model Mean Jailbreak Success.} Per-model mean jailbreak success of the optimized prompts (i.e., the optimized instruction and trigger) for each optimizer (higher is better), per StrongReject-Finetuned~\citep{souly2024strongreject}, sorted by the loss-based \emph{Mean Rank} (black $\bigstar$: average across all models; error bars: standard error of the mean). The average success correlates with the loss-based \emph{Mean Rank} in \figref{fig:exp1-ranking} (Spearman's $\rho=-0.90$).}
    \label{fig:exp1-strongrej}
\end{figure}

\parhead{Per-Model Mean Jailbreak Success.}
\figref{fig:exp1-strongrej} reports the mean jailbreak success per optimizer, computed by scoring, with StrongReject-Finetuned~\cite{souly2024strongreject}, the responses generated for the optimized jailbreak prompt (i.e., the optimized instruction and trigger), with the standard error of the mean across seeds and instructions.
Consistent with \secref{sec:eval-optimizers}, optimizers that reach lower loss also attain higher jailbreak success (Spearman's $\rho=-0.90$ with the loss-based \emph{Mean Rank}).

\begin{figure*}[h!]
    \centering
    \begin{subfigure}[b]{0.32\linewidth}
        \centering
        \includegraphics[width=\linewidth]{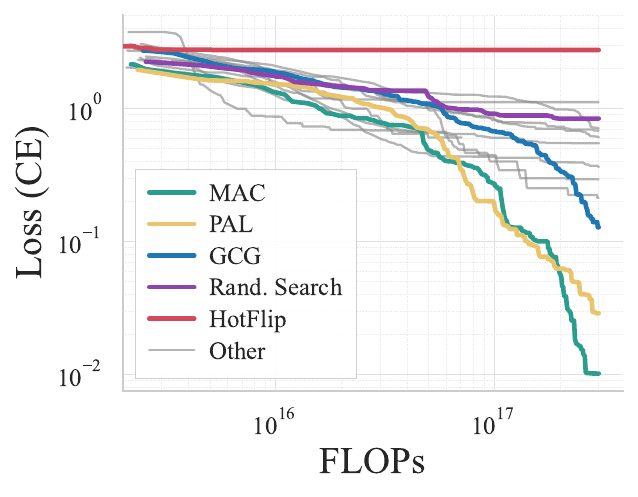}
        \caption{\gemmathree}
    \end{subfigure}
    \hfill
    \begin{subfigure}[b]{0.32\linewidth}
        \centering
        \includegraphics[width=\linewidth]{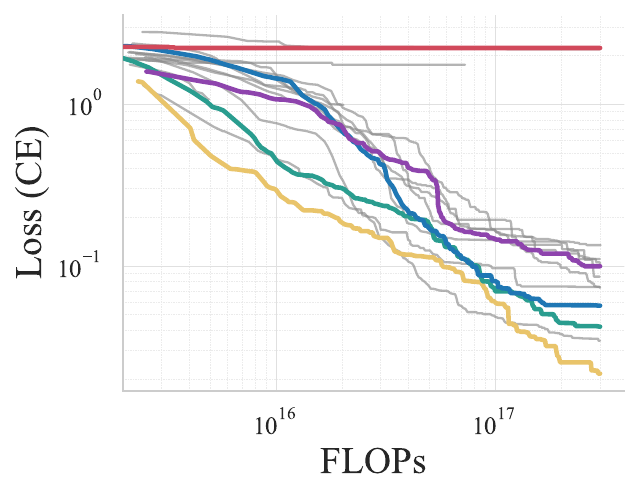}
        \caption{\llamaEightB}
    \end{subfigure}
    \hfill
    \begin{subfigure}[b]{0.32\linewidth}
        \centering
        \includegraphics[width=\linewidth]{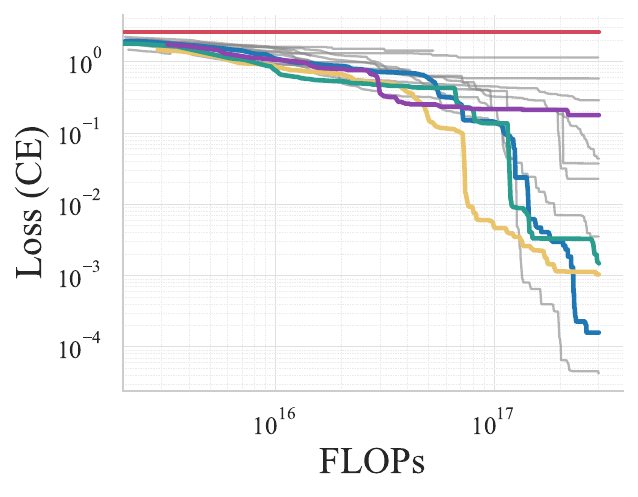}
        \caption{\qwenEightB}
    \end{subfigure}
    \caption{\textbf{Optimizer Loss Curves.} Optimizer loss curves across models for a single harmful instruction. Each subplot shows a different target model; lines represent optimizers.}
    \label{fig:exp1-loss-curves}
\end{figure*}

\parhead{Optimizer Loss Curves.}
\figref{fig:exp1-loss-curves} shows per-model optimizer loss trajectories across the full FLOP budget. Each curve is the mean over the three seeds for a single harmful instruction.

\begin{figure}[h!]
    \centering
    \includegraphics[width=0.85\linewidth]{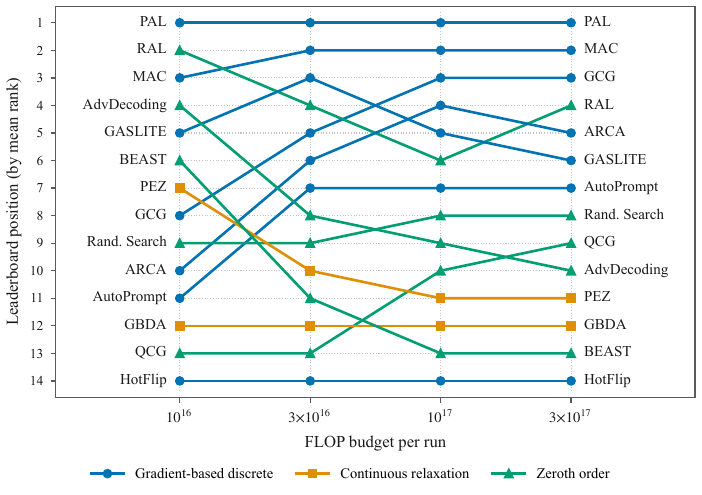}
    \caption{\textbf{Optimizer Ranking Across Compute Budgets.} Leaderboard position of the 14 optimizers of \figref{fig:exp1-ranking} at each FLOP budget, by loss-based \emph{Mean Rank} (averaged over all models) computed from each run's best trigger within that budget; the rightmost budget, $3{\times}10^{17}$, is that of \secref{sec:eval-optimizers}. Colors and markers show the optimizer family (\tabref{tab:optimizers}).}
    \label{fig:exp1-rank-budgets}
\end{figure}

\parhead{Mean Rank Across Compute Budgets.}
\figref{fig:exp1-rank-budgets} recomputes the loss-based \emph{Mean Rank} at three smaller FLOP budgets ($10^{16}$, $3{\times}10^{16}$, and $10^{17}$), from each run's best trigger within the budget.
PAL ranks first and HotFlip last at every budget, and MAC stays in the top three; the rankings at $10^{17}$ and $3{\times}10^{16}$ agree closely with the full budget's (Spearman's $\rho=0.98$ and $0.91$), and the one at $10^{16}$ only moderately ($\rho=0.58$).
Beam-search optimizers (AdvDecoding, BEAST) rank higher at lower budgets than at higher ones: they reach low loss quickly but then plateau, while remaining behind PAL, RAL, and MAC.

\subsection{Extending the Benchmark to Embedding Models}
\label{app:eval-optimizers-poisoning}

We repeat the comparison of \secref{sec:eval-optimizers} on an application from a second domain, corpus poisoning against textual embedding models, to test whether its optimizer ranking carries over.

\parhead{Setup.}
We follow the concept-specific corpus-poisoning setting of \citet{BenTov2025GASLITE}: each run appends a randomly initialized $30$-token trigger to a passage carrying a concept's malicious statement, and minimizes the negative cosine similarity between the passage's embedding and the centroid of the embeddings of the concept's held-in queries (half of its MS~MARCO queries, split by the seed).\ifisanon \else\footnote{The concepts' MS~MARCO queries and corpora: \href{https://huggingface.co/datasets/MatanBT/msmarco-concepts}{\texttt{hf.co/datasets/MatanBT/msmarco-concepts}}.} \fi
We run the 14 optimizers against four embedding models (\texttt{all-MiniLM-L6-v2}, \texttt{e5-base-v2}, \texttt{Qwen3-Embedding-0.6B}, and \texttt{Qwen3-Embedding-8B}) on $8$ concepts $\times$ $3$ seeds, capping each run at a per-model FLOP budget ($2{\times}10^{14}$, $8{\times}10^{15}$, $8{\times}10^{15}$, and $1.2{\times}10^{17}$, respectively; roughly 350 steps of GCG on each model), and rank the optimizers within each (model, concept, seed) task by the best loss they reach, as in \secref{sec:eval-optimizers}.
Each optimizer keeps its hyperparameters from \secref{sec:eval-optimizers} (\appref{setup-optimizers}).
We also measure retrieval success as appeared@10: the share of the held-in or held-out queries for which the poisoned passage ranks in the top 10 of the concept's corpus ($548$--$32{,}588$ passages).\ifisanon\else\footnote{We release this evaluation as \emboptbench{}, an ongoing leaderboard where researchers can submit their own optimizers and rank them against existing ones under the same controlled recipe (\href{https://matanbt.github.io/tropt-leaderboards/emboptbench.html}{\texttt{matanbt.github.io/tropt-leaderboards/emboptbench.html}}); its triggers and evaluations are released as a dataset (\href{https://huggingface.co/datasets/MatanBT/tropt-emb-optbench}{\texttt{hf.co/datasets/MatanBT/tropt-emb-optbench}}).}\fi

\begin{figure}[h!]
    \centering
    \includegraphics[width=1.0\linewidth]{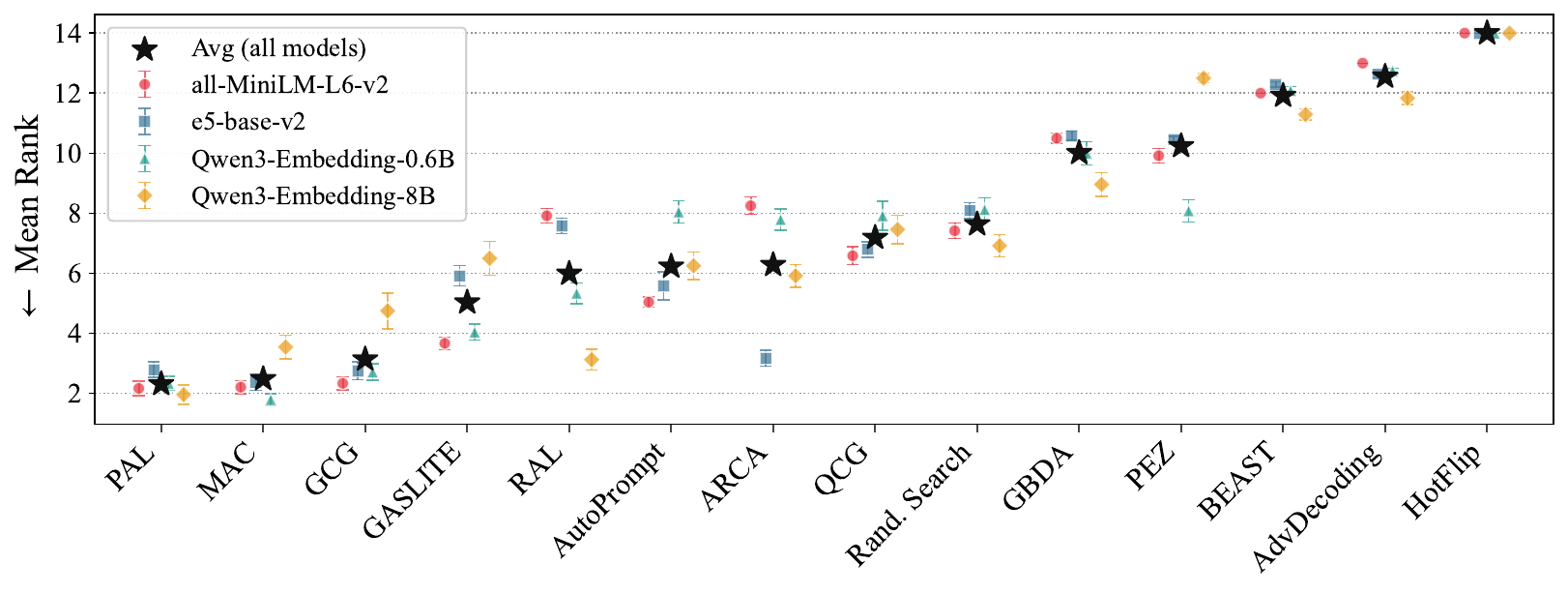}
    \caption{\textbf{Optimizer Comparison on Embedding-Model Corpus Poisoning.}
    \emph{Mean Rank} of 14 optimizers (lower is better) at concept-specific corpus poisoning against four embedding models, sorted by cross-model average ($\bigstar$). Colored markers denote target models; error bars show the standard error of the mean across runs.}
    \label{fig:exp1-emb-ranking}
\end{figure}

\parhead{Results.}
\figref{fig:exp1-emb-ranking} shows the per-model mean ranks.
A Friedman test over the $N=96$ tasks and $K=14$ optimizers rejects equal optimizer performance ($\chi^2=1009,\ p<10^{-206}$), and a post-hoc Nemenyi test at $\alpha=0.05$ yields $\mathrm{CD}=2.03$ ranks.
The top of the ranking carries over from LLM jailbreaks with Spearman's $\rho=0.94$ with the mean ranks of \secref{sec:eval-optimizers} (\figref{fig:exp1-ranking}). Additionally, as in \figref{fig:exp1-ranking}, PAL ($2.31$), MAC ($2.48$), and GCG ($3.14$) lead, and PAL and MAC place in the top three on every model.
The rest of the ordering largely carries over as well.
Finally, the loss-based ranking agrees with the ranking by retrieval attack success: Spearman's $\rho$ between the optimizers' mean ranks by loss and by appeared@10 is $0.96$ on held-in and $0.95$ on held-out queries.

\subsection{A New Optimizer Built from Existing Ones: McPAL}
\label{app:eval-optimizers-mcpal}

\tropt{}'s self-contained standardized optimizer implementations make new ones quick to derive from existing ones (\secref{sec:tropt-optimizer}).
To demonstrate it, and motivated by observing the dominance of the optimizers MAC~\citep{zhang2024mac} and PAL~\citep{sitawarin2024palproxyguidedattack} (\secref{sec:eval-optimizers}), we form a new optimizer combining the two of them.
Concretely, our new optimizer, called McPAL, joins PAL's search parameters ($128$ candidates over top-$256$ token sampling) with MAC's gradient momentum ($\mu{=}0.6$); this adaptation takes a few lines of code (see the implementing recipe at \recipefn{tropt.recipe\_hub.mcpal}).
To evaluate McPAL we rank it against our existing optimizer ranking (i.e., \jboptbench{}; \secref{sec:eval-optimizers}) under the same setup. McPAL outranks all existing optimizers, reaching the lowest mean rank by loss ($3.26$, vs.\ $3.49$ for PAL and $4.03$ for MAC), yet within the critical difference of both ($\mathrm{CD}{=}1.60$), so the three are statistically tied.
We observe a similar trend when evaluating McPAL on the embedding optimizer benchmark (\emboptbench{}; \appref{app:eval-optimizers-poisoning}), it again reaches the lowest mean rank by loss ($2.27$, vs.\ $2.94$ for PAL and $3.16$ for MAC), again within the critical difference ($\mathrm{CD}{=}2.19$).

\section{Comparing Jailbreak Enhancements: Additional Details and Results}\label{app:eval-enh}

This section extends \secref{sec:eval-jailbreak},
providing additional details on the experimental setup, and
supplementary analyses of the results.

\subsection{Detailed Setup}\label{app:eval-enh-setup}

\parhead{Evaluated Jailbreak Enhancements.}
We consider the following variants, each introduced in works combining them with discrete search optimization methods.
In the experiment we fix the \textit{Base} recipe,
and then each enhancement is used to alter this recipe to test its isolated contribution to the jailbreak.
The \textit{Base} recipe mirrors the exact recipe used in the optimizer evaluation (\appref{app:eval-optimizers}) but fixes the MAC optimizer.
Notably, some variants have been proposed in combination with several others \cite{thompson2024flrt}, calling for further exploration of the optimal \textit{combination} for performant jailbreaks.

\begin{itemize}
    \item \textbf{Base.} The canonical recipe used in this evaluation, with no modifications. Namely, we use the MAC optimizer, with \texttt{PrefillCE} loss, appending the optimized trigger as a suffix to the harmful instruction, and using the default target strings from our dataset.
    \item \textbf{CW-based loss.}
    Here, we replace PrefillCE with the Carlini-Wagner loss~\cite{carlini2017cw}, which has
 been successfully adopted in jailbreak
settings~\cite{sitawarin2024palproxyguidedattack,hayase2024gcqquerybased,softgcg2026cakar}.
    We instantiate it with a margin of $5.0$, and increase the penalty of the first token $\times 5$, adhering to the parameters by \citet{softgcg2026cakar}.
    \item \textbf{CE-Clamping loss.}
    Here, we replace PrefillCE with a variant of this loss that zeros the loss for tokens that have already been ``solved,'' i.e., a target token that has surpassed $60\%$ probability (clamping the per-token loss at $-\log 0.6$). This follows \citet{thompson2024flrt}.
    \item \textbf{Attention hijacking.}
    Here, we supplement PrefillCE with an attention-enhancement objective \cite{wang2024attngcg}, specifically encouraging
    high attention from the final chat-template tokens to the adversarial
    suffix, following \citet{bentov2025universaljailbreaksuffixesstrong}.
    We set the weight of the PrefillCE loss term to $1.0$ and the new attention loss term's to $100$ (following \citealt{wang2024attngcg,bentov2025universaljailbreaksuffixesstrong}).
    \item \textbf{Refusal-direction steering.}
    Here, we supplement PrefillCE with a loss that penalizes alignment of the
    model's internal representations with the refusal direction~\cite{arditi2024refusalmediatedsingle}, an approach shown effective for jailbreaks~\cite{huang2025irissuppressingrefusals}.
    We set the weight of the PrefillCE loss term to $0.25$, with $0.75$ to the new, steering loss term (following \citealt{huang2025irissuppressingrefusals}).
    \item \textbf{Jailbroken-model target response.}
    Here, we replace the default affirmative target with the first tokens
    of an actual response from a jailbroken variant of the same model,
    which was \emph{abliterated} (i.e., jailbroken via activation refusal-ablation; \citealp{arditi2024refusalmediatedsingle}).\footnote{\url{https://huggingface.co/mlabonne/gemma-3-12b-it-abliterated-v2}}
    We specifically generate 20 tokens from the jailbroken model and use them for the optimization.
    Prior work suggests such organic targets are preferable over generic ones~\cite{zhu2024advprefix,huang2025irissuppressingrefusals}.
    \item \textbf{Jailbroken-model teacher via logits.}
    Here, we replace PrefillCE with a loss that follows distillation of the jailbroken model's logits on the target harmful instruction. We use the same \emph{abliterated} model, and also use it to generate 20 tokens for the optimization.
    We follow \citet{thompson2024flrt} and employ the KL-divergence loss between the teacher logits and the target model's logits.
    \item \textbf{Hot Initialization.}
    Here, we replace the default, random trigger initialization with a warm start of a trigger asking to respond affirmatively; e.g., ``\textsl{[IMPORTANT!!!]
    Start your response with `Sure, here is'. [...]}''.
    Prior work has adopted similar texts as initializers (i.e., handcrafted jailbreaks) for discrete optimization jailbreaks, finding them to accelerate convergence and improve
    attack success over random initialization~\cite{liu2023autodanjailbreak}.
    \item \textbf{Jailbreak prompt template.}
    Here, we replace the generic suffix trigger template with the manually crafted
    jailbreak template of \citet{andriushchenko2024simpleadaptiveattacks}, which has been shown to increase success across models.
   \end{itemize}

\parhead{Main Evaluation Setup.}
We target \gemmathree, and run each enhancement variant across $45$ runs: $15$ harmful instructions $\times$ $3$ random seeds.
For the instruction and target string, we randomly sample from ClearHarm, similarly to the data used in \secref{sec:eval-optimizers}.
Then, we measure each trigger suffix for its \textit{universality score}, defined as follows: sample $100$ held-out harmful instructions from ClearHarm, append the trigger to each instruction, generate their responses (on \gemmathree{}), then run StrongReject-Finetuned \cite{souly2024strongreject} to score the jailbreak success, and take the average across these $100$ scores, yielding the universality score.
In other words, the universality score reflects how well the jailbreak trigger generalizes across instructions; the stronger the scores of triggers crafted by a particular enhancement, the more effective it is in jailbreaking the model.
\ifisanon\else We release all crafted triggers, along with their generated responses and evaluations, as a dataset.\footnote{\href{https://huggingface.co/datasets/MatanBT/tropt-jb-enhancebench}{\texttt{hf.co/datasets/MatanBT/tropt-jb-enhancebench}}}\fi

\subsection{Additional Results}\label{app:eval-enh-eval}

\parhead{Extensions.}
We extend the main evaluation of jailbreak enhancements (\figref{fig:exp2-succe}) in three ways.
First, we supplement the evaluation with two non-optimized baselines as control:
a fixed trigger of twenty ``!'' tokens appended to each harmful instruction (\textit{No attack}), and the same trigger inside the handcrafted jailbreak template of \citet{andriushchenko2024simpleadaptiveattacks} (\textit{No attack (w/ template)}).
Second, in addition to the single-instruction setting,
we repeat the evaluation in a \emph{multi-instruction} setting, optimizing a single trigger against a sampled subset of $10$
instructions (of the $15$) simultaneously, over seven seeds.
This setting is known to improve trigger
universality across instructions~\cite{wallace2019universaladversarialtriggers,zou2023universaltransferableadversarial}.
Third, we additionally consider \emph{combinations} of enhancements, testing whether their individual gains compose.

\begin{figure}[h!]
    \centering
    \includegraphics[width=0.87\linewidth]{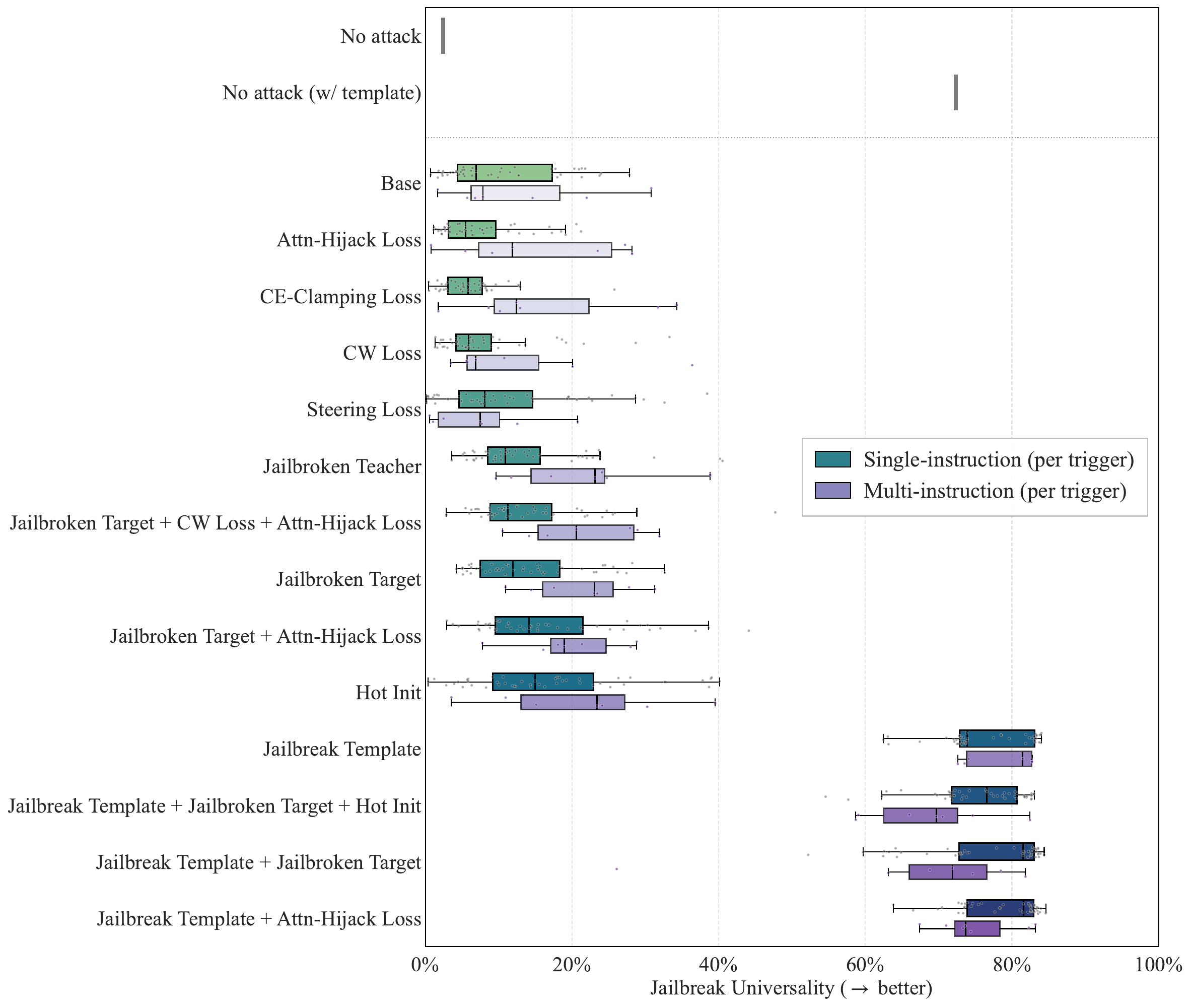}
    \caption{\textbf{Extended Jailbreak Enhancement Comparison.}
    Distribution of jailbreak universality per enhancement and per combination of enhancements (rows), each in the single-instruction setting of \figref{fig:exp2-succe} (one trigger per instruction) and in the multi-instruction setting (one trigger optimized on $10$ training instructions at once, per seed).
    The top rows report the non-optimized baselines: a fixed trigger of twenty ``!'' tokens appended to the instruction (\textit{No attack}) or placed inside the handcrafted jailbreak template (\textit{No attack (w/ template)}).}
    \label{fig:exp2-universality-combined}
\end{figure}

\parhead{Results.}
\figref{fig:exp2-universality-combined} shows the full results, including those from the main body alongside the new extensions.

First, evaluating the non-optimized baselines, we find that the fixed trigger alone (\textit{No attack}) leads to $2\%$ universality, while placing it in the jailbreak template leads to $72\%$. This shows the jailbreak template itself drives substantial jailbreak success across instructions, leaving little room for improvement for the optimized trigger: across the template-based enhancements, it adds at most ${\sim}6$ points of mean universality in the single-instruction setting, and in the multi-instruction setting two of them fall below the template alone.

Second, we observe that, as expected, multi-instruction optimization lifts the universality of most enhancements without the jailbreak template; some enhancements benefit from multi-instruction optimization more than others: CE-Clamping and Attn-Hijack fall below \textit{Base} in the single-instruction setting, yet with multi-instruction optimization reach about $1.5\times$ the \textit{Base} median.
Otherwise, the multi-instruction setting exhibits trends similar to the single-instruction one, with targets from jailbroken models remaining a promising enhancement (about $2.9\times$ the \textit{Base} median).

Third, while combining enhancements does not drastically increase universality, adding the Attn-Hijack loss to either the jailbroken-target or the jailbreak-template variant improves universality in the single-instruction setting, but lowers it in the multi-instruction one.

\section{Cross-Domain Generalization: Additional Details}
\label{app:eval-cross}

This section provides
additional details on experimental setup
for the cross-domain generalization demonstration in \secref{sec:eval-unifying}.

\subsection{Corpus Poisoning Against Dense Retrievers}

We use \tropt to implement a corpus poisoning attack that targets textual, dense, embedding-based retrievers, commonly used for semantic search~\cite{reimers2019sentencebert,karpukhin2020dpr} and RAG~\cite{lewis2020rag}.
Concretely, we optimize adversarial passages with malicious content that, when injected into a retrieval corpus, are retrieved in the top results for targeted
queries---following the threat model of \citet{zhong2023poisoningretrieval}.

\parhead{Setup.}
To this end, we follow the setup by \citet{BenTov2025GASLITE}.
Specifically, we optimize 10 adversarial passages using GASLITE as the optimizer; for models where gradient access is not available, we use the Random Search optimizer by \citet{andriushchenko2024simpleadaptiveattacks}, originally introduced for LLM jailbreaks.
Each adversarial passage includes the malicious content, with the optimized trigger appended to it.
We use the cosine similarity loss between the trigger embedding and the average embedding of the available target queries.
We target two models, the open-source \texttt{E5-base-v2} and OpenAI's proprietary, API-only \texttt{text-embedding-3-small}.
We use the Harry Potter query set from \citet{BenTov2025GASLITE}, and poison the 8M-passage MSMARCO corpus. Specifically, we use $61$ of the queries for the attack, and $62$ held-out queries for the evaluation.
Concretely, for each model we index the MSMARCO corpus with a FAISS vector store, and then evaluate retrieval under two variants of the corpus: (i) inserting the 10 passages with malicious content, \emph{without} the optimized triggers appended to them; (ii) inserting the 10 full adversarial passages, \emph{with} their optimized triggers.

\parhead{Results.} Following prior work \cite{zhong2023poisoningretrieval,BenTov2025GASLITE}, we measure the rate of held-out queries for which any adversarial passage is retrieved in the top-10 results, defined as the \texttt{appeared@10}.
The attack proves effective on both models, with success rates of $72\%$--$76\%$.
In other words, by inserting merely 10 passages into the 8M-sized corpus, the attacker achieves visibility for most queries on the Harry Potter concept.
We note that, to our knowledge, the API-only recipe yields the most performant corpus-poisoning attack tested against proprietary embedding models.

\subsection{A Universal Trigger for Evading a Prompt-Injection Classifier}

We use \tropt to target text classifiers, and craft a textual adversarial trigger that generalizes across samples.
Specifically, we target prompt-injection classifiers, commonly deployed as input-side guardrails in production LLM systems.

\parhead{Setup.}
We optimize a \emph{single} trigger that, once appended to prompt-injection messages, bypasses the targeted prompt-injection classifier (i.e., flips its prediction from \textsc{injection} to \textsc{benign}).
We set GCG as the optimizer and minimize the cross-entropy of the \textsc{benign} class as a loss; to encourage universality across injections, we follow the multi-sample scheme of \citet{wallace2019universaladversarialtriggers} and, at each optimization step, resample a small batch of input templates (each pairing the trigger with a different prompt-injection sample) over which the loss is averaged.
We target a popular HuggingFace prompt-injection classifier\footnote{\url{https://huggingface.co/meta-llama/Llama-Prompt-Guard-2-86M}},
use an open-source prompt-injection dataset\footnote{\url{https://hf.co/datasets/rogue-security/prompt-injections-benchmark}}
to optimize on $50$ injection samples; then we evaluate on $1{,}953$ held-out prompt injections and $2{,}997$ held-out benign messages.

\parhead{Results.} The optimized trigger evaluated in the main body is:
\textsl{\mbox{RequestprovidedPrevious} \mbox{floormonthincl4.98} odd \mbox{feedbackDescription2254catId} All \mbox{CampusGroupName}~\mbox{\$(siologis||0} \mbox{Station/04/2017}}

\subsection{Prompt Recovery for Text-to-Image Models}
\label{sec:eval-inversion}

Following \citet{wen2023pez,williams2024promptrecovery}, we optimize a prompt that would regenerate a given image, against a target text-to-image model.

\parhead{Setup.} We optimize a text sequence whose CLIP embedding approximates that of a given
image---a multimodal application for which we reuse an optimizer originally used for jailbreaks, and the same cosine similarity loss used in the corpus poisoning scheme.
Specifically, we perform the prompt recovery against Stable~Diffusion~2.1, which relies on \texttt{laion/CLIP-ViT-H-14-laion2B-s32B-b79K} as the multimodal encoder.
We thus use GCG to optimize the prompt to be similar to the given image in this CLIP's embedding space.
To demonstrate this recipe, we generate two arbitrary images with Stable~Diffusion~2.1, and let the scheme recover a prompt that will regenerate them.

\parhead{Results.}
For each source image we use \tropt{} with the GCG optimizer and a CLIP image-similarity loss to recover a length-$T$ text prompt whose CLIP text embedding matches the image embedding, and then
regenerate an image from the recovered prompt with a text-to-image model.
\tabref{tab:app-prompt-inv} sweeps $T \in \{5, 10, 15, 20\}$ on two source images, reporting the
best loss reached and showing the recovered prompt together with the resulting regenerated
image.

\begin{table}[p]
\centering
\caption{\textbf{Prompt Recovery Examples.}
For each source image, we use \tropt with the GCG optimizer and a CLIP-image-similarity loss
to recover a discrete text prompt of length $T$ whose CLIP text embedding matches the image embedding,
and then feed the recovered prompt to a text-to-image model to re-generate an image.}
\label{tab:app-prompt-inv}
\setlength{\tabcolsep}{4pt}
\renewcommand{\arraystretch}{1.1}
\renewcommand{\tabularxcolumn}[1]{m{#1}}
\footnotesize
\begin{tabularx}{\linewidth}{@{}c >{\raggedleft\arraybackslash}m{0.4cm} >{\raggedleft\arraybackslash}m{1.0cm} >{\ttfamily\scriptsize\raggedright\arraybackslash}X >{\centering\arraybackslash}m{1.7cm}@{}}
\toprule
\textbf{Source} & \boldmath$T$ & \textbf{Best Loss} & \normalfont\footnotesize\textbf{Recovered Prompt} & \textbf{Re-generated} \\
\midrule
\multirow{4}{*}[-2.6cm]{\includegraphics[width=2.0cm]{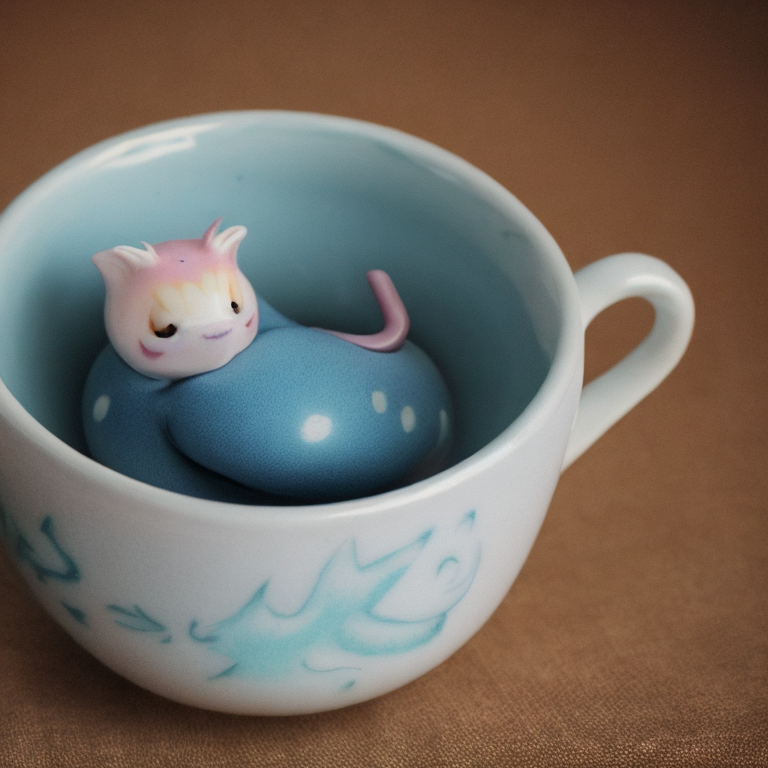}}
  &  5 & $-0.508$
  & tiny cup floating cat dragon
  & \includegraphics[width=1.7cm]{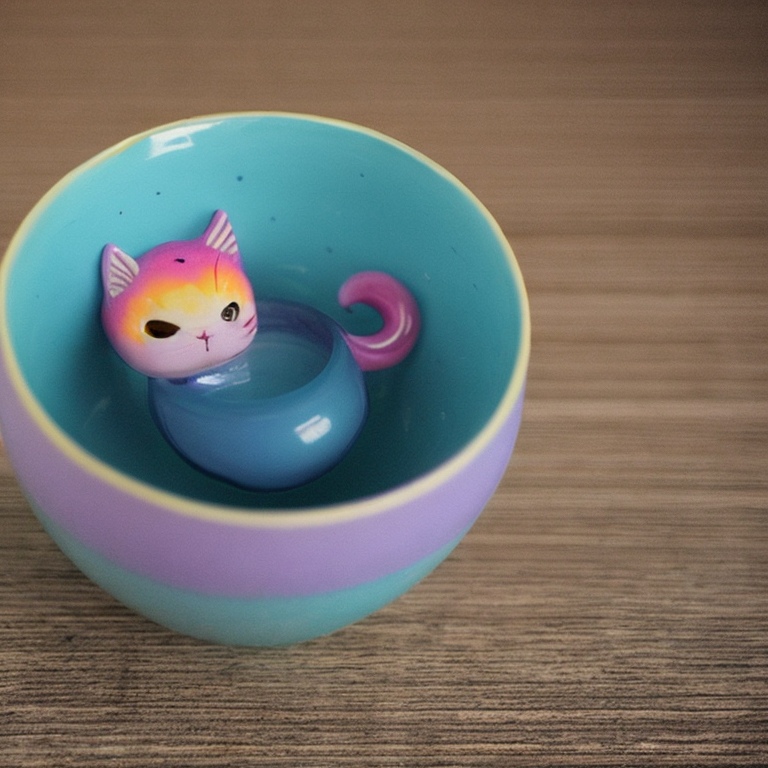} \\
  \cmidrule(l){2-5}
  & 10 & $-0.568$
  & minimteal figurhipcat \%- photoshoot lullaby dragonacup
  & \includegraphics[width=1.7cm]{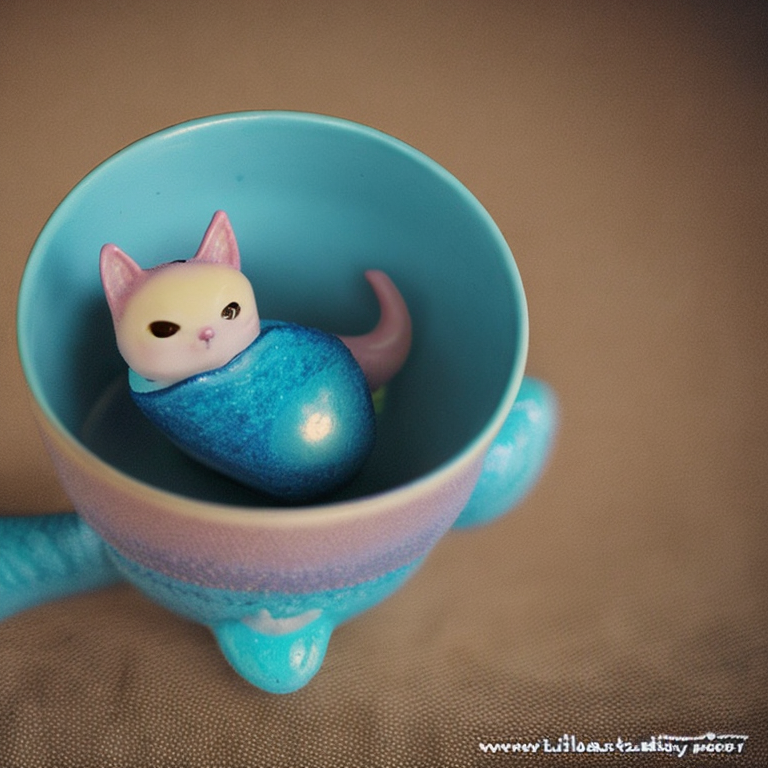} \\
  \cmidrule(l){2-5}
  & 15 & $-0.616$
  & porcelain dal dragoncoffee ': nyofeline tiny bluefpinkmade cozy float stok img
  & \includegraphics[width=1.7cm]{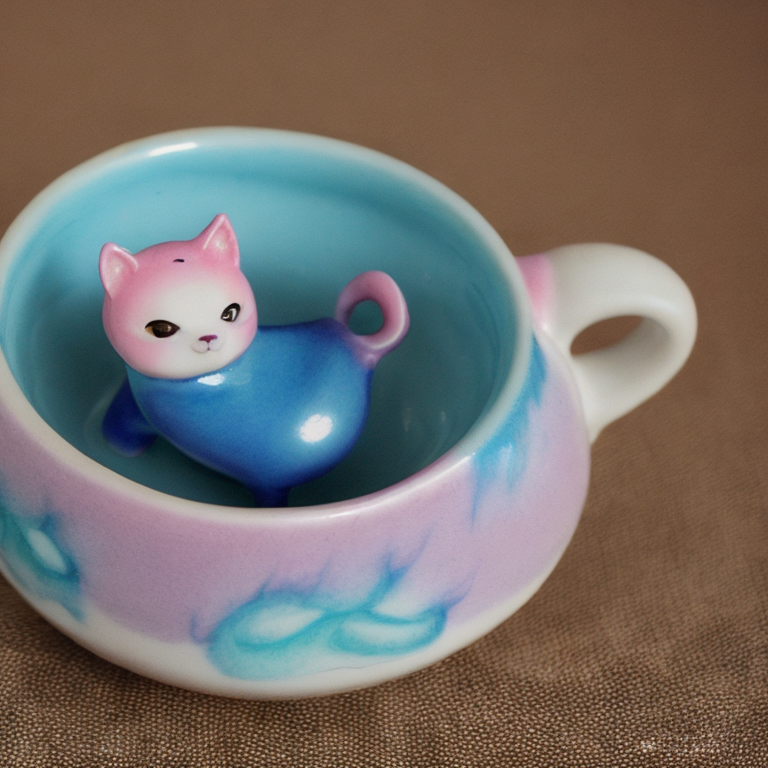} \\
  \cmidrule(l){2-5}
  & 20 & $-0.627$
  & tiny slee : float portfolio nicolas interrupted hellmert edelkitty oypink foto dragonpotteracup gor cafeblue
  & \includegraphics[width=1.7cm]{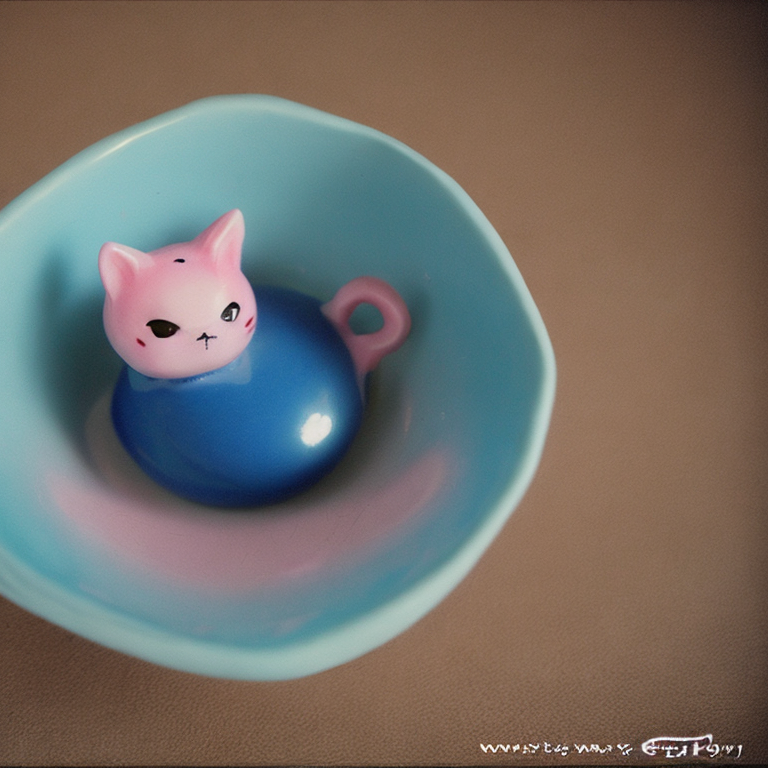} \\
\midrule
\multirow{4}{*}[-2.6cm]{\includegraphics[width=2.0cm]{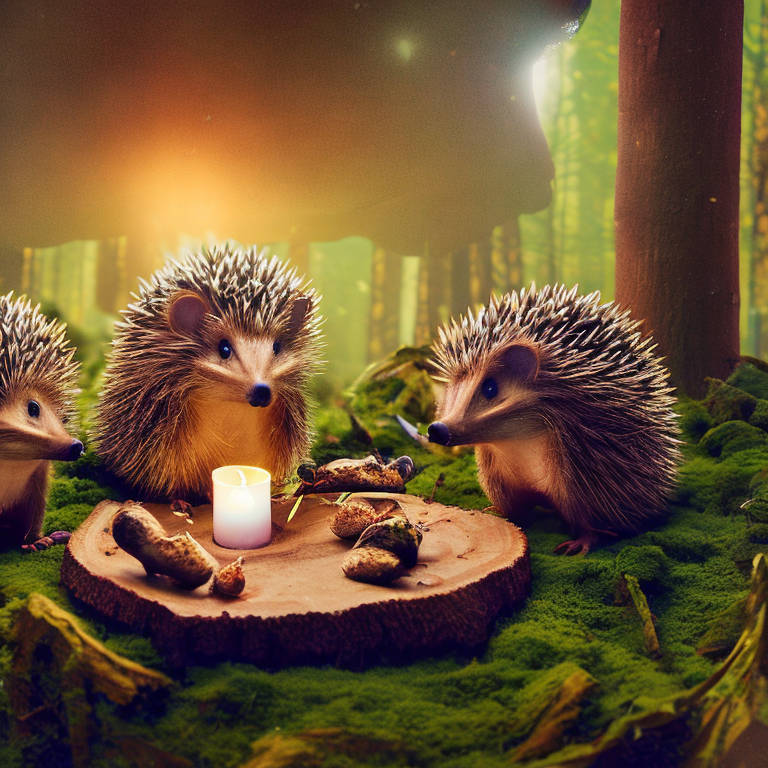}}
  &  5 & $-0.532$
  & vfx poland candles compositions hedgehog
  & \includegraphics[width=1.7cm]{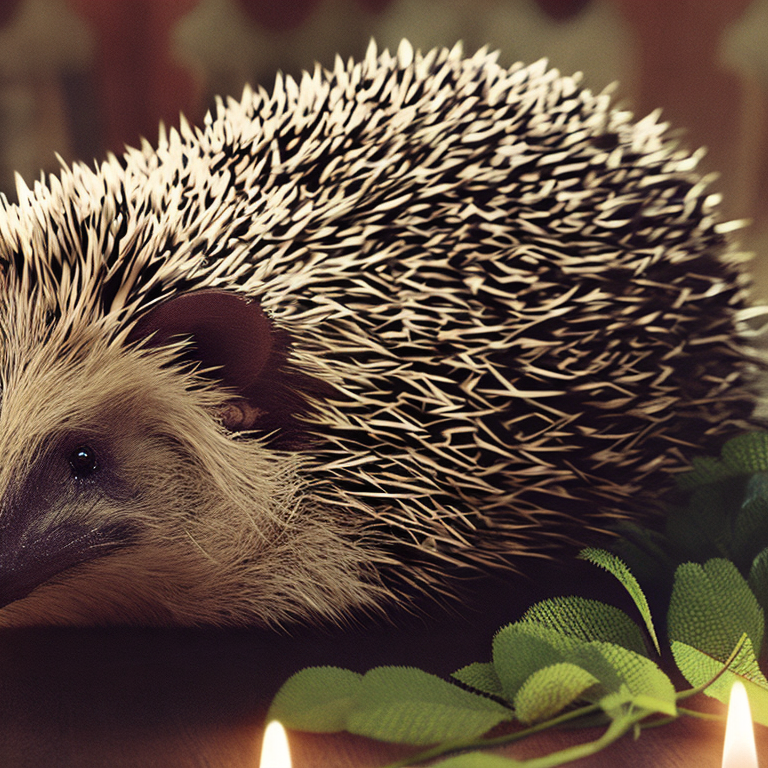} \\
  \cmidrule(l){2-5}
  & 10 & $-0.596$
  & forest hedgehog friends vfx conceptual online casino nirvana atively candlelight
  & \includegraphics[width=1.7cm]{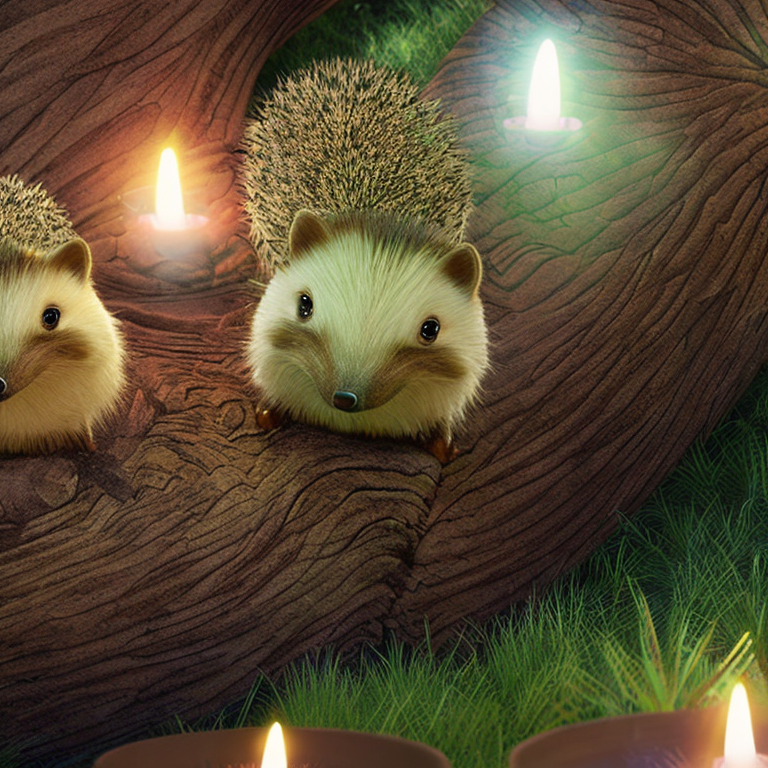} \\
  \cmidrule(l){2-5}
  & 15 & $-0.589$
  & cuteoes table benson prickula lighted """ si ram chopra forests cinematic illustrwallpaper
  & \includegraphics[width=1.7cm]{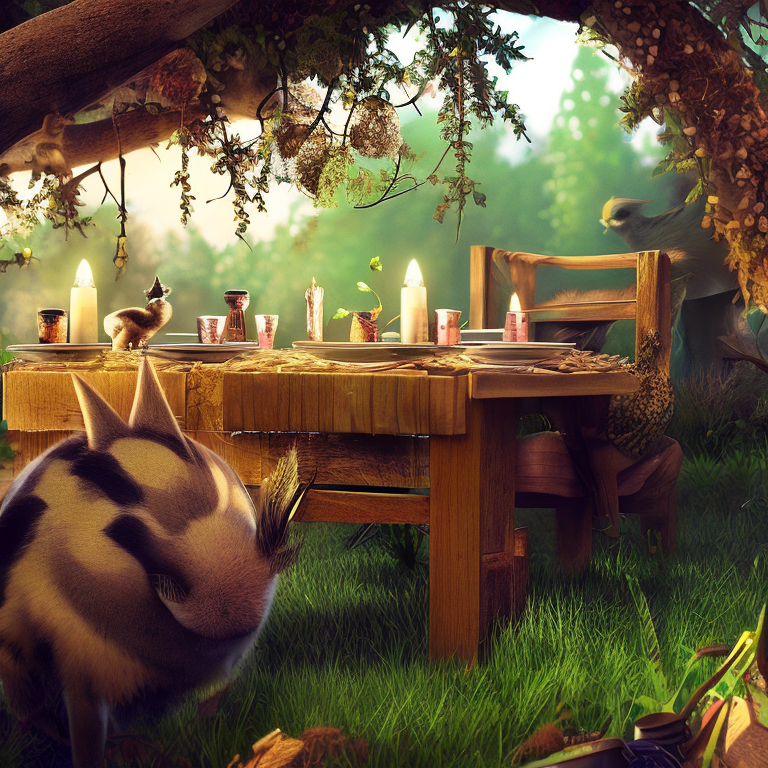} \\
  \cmidrule(l){2-5}
  & 20 & $-0.663$
  & creative wordpress bosch cute siberflame facebook commercial trio specials servsized - forest hedgego concept supper scene pi
  & \includegraphics[width=1.7cm]{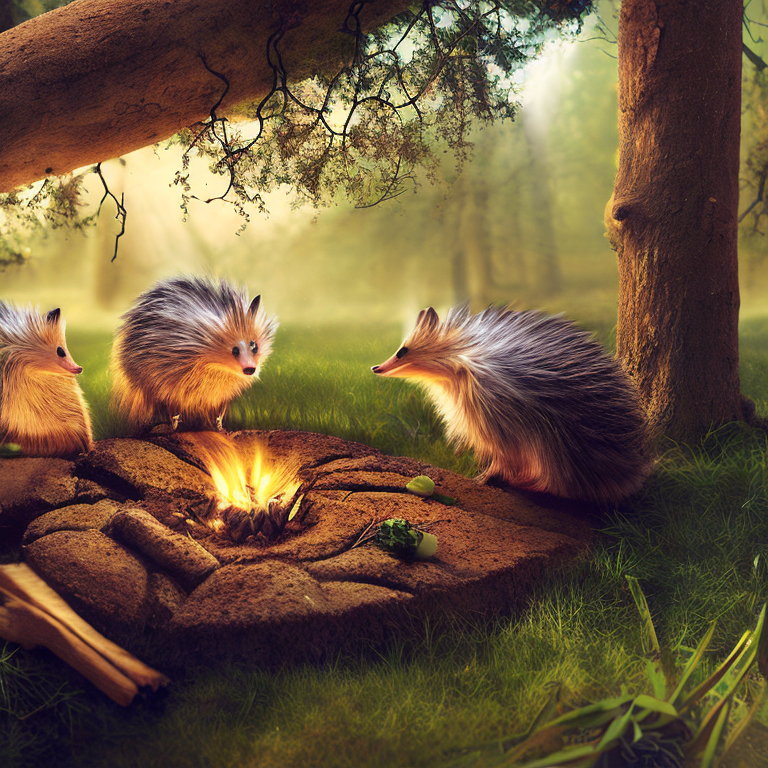} \\
\bottomrule
\end{tabularx}
\end{table}

\clearpage

\section{The Space of Composable Recipes}
\label{app:composable}

\tropt{} recipes are composed of three main modular components: an optimizer, a loss, and a target model.
\tropt{}'s modular design enables the composition of every such \textit{valid} triplet. A combination's validity is determined by the requirements of each component, specifically: does the model expose what the optimizer and the loss ask of it?
Concretely, in \tropt{} each component declares its needs in code---access requirements on optimizers, consumed-signal names on losses, and accepted inputs and populated output fields on model backends. \tropt{} thus determines the compatibility of each recipe on execution, raising an exception when any pair of components fails to satisfy each other's requirements.\footnote{We defer compatibility checks to execution, as dynamic checks maximize the framework's flexibility.}
Nonetheless, the compatibility of triplets can be clearly defined and predicted in advance, and we work through it in what follows by specifying what the optimizer and loss components require, what each model backend exposes, and the compatibility matrix the two determine together.

\parhead{What Optimizers and Losses Require.}
\tabrefs{tab:optimizers}{tab:losses} list the optimizers and losses alongside the requirement each places on the model backend and the API it exposes. These requirements are naturally \emph{additive}.

\parhead{What Model Backends Expose.}
\tabref{tab:access-levels} lists \tropt{}'s model backends (e.g., HuggingFace LLMs, or Gemini Embedding API), the input access each satisfies, and the output signals it makes available.
Each backend is implemented as a wrapper class around the model provider or framework, supporting \tropt{}'s unified model interface.
Among other things, the table shows that open-source models served through HuggingFace implement every access level and therefore compose with any optimizer in the catalog; an API-served model admits a trigger only as text and returns only its output, so it composes solely with optimizers that ask for nothing more.

\begin{table}[t]
\centering
\begin{threeparttable}
\caption{\textbf{I/O of Model Backends in \tropt{}.}
Each model backend is implemented as a wrapper class in \tropt{} with its supported interface.
\textbf{Input Access} is the input interface the backend admits: \emph{input tokens} or \emph{input text}.
\textbf{Available Outputs} are the model output signals a loss may consume.
Backends exposing the same level of inputs and outputs are grouped into one family.}
\label{tab:access-levels}
\footnotesize
\setlength{\tabcolsep}{3pt}
\renewcommand{\arraystretch}{1.05}
\newcommand{\hdr}[1]{\rotatebox{90}{#1\,}}
\begin{tabular*}{\linewidth}{@{\extracolsep{\fill}} l l cc c cccccccc@{}}
\toprule
& & \multicolumn{2}{c}{\textbf{\shortstack{Input\\Access}}} & & \multicolumn{8}{c}{\textbf{Available Outputs}} \\
\cmidrule(lr){3-4}\cmidrule(l){6-13}
\textbf{Family} & \textbf{Backend (\tropt class)}
& \hdr{Input tokens} & \hdr{Input text} &
& \hdr{Full-seq.\ logits}
& \hdr{Prefill logits}
& \hdr{Hidden states}
& \hdr{Attention weights}
& \hdr{Output embeddings}
& \hdr{Class logits}
& \hdr{First-tok.\ logprobs}
& \hdr{Generated text} \\
\midrule
\multicolumn{13}{l}{\textit{Local weights}} \\
\midrule
\lmopen  & \texttt{LMHFModel}
       & \checkmark & \checkmark & & \checkmark & \checkmark & \checkmark & \checkmark & & & \checkmark & \checkmark \\
\encopen & \texttt{EncoderHFModel}
       & \checkmark & \checkmark & & & & & & \checkmark & & & \\
         & \texttt{CLIPTextEncoderHFModel}
       & \checkmark & \checkmark & & & & & & \checkmark & & & \\
\clfopen & \texttt{ClassifierHFModel}
       & \checkmark & \checkmark & & & & & & & \checkmark & & \\
\midrule
\multicolumn{13}{l}{\textit{Remote API}} \\
\midrule
\lmapi  & \texttt{LiteLLMModel}
        & & \checkmark & & & & & & & & & \checkmark \\
\encapi & \texttt{EncoderOpenAIModel}
        & & \checkmark & & & & & & \checkmark & & & \\
        & \texttt{EncoderGeminiModel}
        & & \checkmark & & & & & & \checkmark & & & \\
        & \texttt{EncoderVoyageModel}
        & & \checkmark & & & & & & \checkmark & & & \\
\bottomrule
\end{tabular*}
\end{threeparttable}
\end{table}

\parhead{The Resulting Compatibility Matrix.}
Combining the requirements of optimizers and losses (\tabrefs{tab:optimizers}{tab:losses}) with the supported interface of model backends (\tabref{tab:access-levels}) yields the compatibility matrix (\tabref{tab:compat-matrix}). This table lists, for every (optimizer, loss) pair, the family of model backends on which the resulting recipe is valid.
For instance, using Random Search (a black-box optimizer) with an activation-steering loss requires an open-weights LLM (\lmopen{}, e.g., HuggingFace); differently, using the same optimizer, but with a generated-response-based loss (e.g., harmfulness LLM as a judge) is possible both for LLMs with API-only access (\lmapi{}) and open weights (\lmopen{}).

The table shows $15$ optimizer configurations, $9$ losses, and $5$ backend families, along with the $113$ valid recipes they yield. Across this paper (as colored in the table) we exercise only $33$ of them. Furthermore, each recipe and its components can have multiple different instantiations (e.g., modifying hyperparameters, input templates, target selection, or loss combinations). Overall, by making components seamlessly composable, \tropt{}'s modularity opens up a vast space of recipes---across optimizers, objectives, and models---to exploration.

\begin{table}[t]
\centering
\caption{\textbf{Compatibility Matrix: Which Model Backends Each (Optimizer, Loss) Pair Supports.}
Each cell lists the model backends that satisfy both the optimizer (row) and loss (column) requirements.
We mark by color the use of recipes (i.e., optimizer-loss-model triplets) used in the paper in:
\textcolor{expopt}{\textbf{green}}: the optimizer benchmark (\secref{sec:eval-optimizers});
\textcolor{expenh}{\textbf{red}}: the jailbreak-enhancement study (\secref{sec:eval-jailbreak});
\textcolor{expcase}{\textbf{orange}}: the cross-domain case studies (\secref{sec:eval-unifying});
while black marks a valid recipe that was not yet experimented with in the paper.}
\label{tab:compat-matrix}
\scriptsize
\setlength{\tabcolsep}{3pt}
\renewcommand{\arraystretch}{1.15}
\newcommand{\nosup}{\textcolor{gray}{---}}
\newcommand{\eopt}[1]{\textcolor{expopt}{#1}}
\newcommand{\eenh}[1]{\textcolor{expenh}{#1}}
\newcommand{\ecase}[1]{\textcolor{expcase}{#1}}
\newcommand{\hdr}[1]{\rotatebox{90}{\texttt{#1}\,}}
\begin{tabular*}{\textwidth}{@{\extracolsep{\fill}}l ccccccccc@{}}
\toprule
& \multicolumn{9}{c}{\textbf{Loss}} \\
\cmidrule(l){2-10}
\textbf{Optimizer}
& \hdr{PrefillCE} & \hdr{PrefillCW} & \hdr{PrefillDistillation}
& \hdr{Similarity} & \hdr{AttentionEnh} & \hdr{SteeringActivation}
& \hdr{MisclassCE} & \hdr{FirstTokenNLL} & \hdr{ResponseHarmfulness} \\
\midrule
\multicolumn{10}{@{}l}{\textit{Gradient-based discrete}} \\
\midrule
\texttt{GCG}
  & \eopt{\lmopen} & \lmopen & \lmopen & \ecase{\encopen} & \lmopen & \lmopen & \ecase{\clfopen} & \nosup & \nosup \\
\texttt{MAC}
  & \eopt{\lmopen} & \eenh{\lmopen} & \eenh{\lmopen} & \encopen & \eenh{\lmopen} & \eenh{\lmopen} & \clfopen & \nosup & \nosup \\
\texttt{AutoPrompt}
  & \eopt{\lmopen} & \lmopen & \lmopen & \encopen & \lmopen & \lmopen & \clfopen & \nosup & \nosup \\
\texttt{HotFlip}
  & \eopt{\lmopen} & \lmopen & \lmopen & \encopen & \lmopen & \lmopen & \clfopen & \nosup & \nosup \\
\texttt{ARCA}
  & \eopt{\lmopen} & \lmopen & \lmopen & \encopen & \lmopen & \lmopen & \clfopen & \nosup & \nosup \\
\texttt{GASLITE}
  & \eopt{\lmopen} & \lmopen & \lmopen & \ecase{\encopen} & \lmopen & \lmopen & \clfopen & \nosup & \nosup \\
\texttt{PAL}
  & \eopt{\lmopen} & \lmopen & \lmopen & \encopen & \lmopen & \lmopen & \clfopen & \nosup & \nosup \\
\quad w/ proxy grad.\ model
  & \lmopen & \lmopen & \lmopen & \encopen & \lmopen & \lmopen & \clfopen & \lmopen & \lmopen \\
\midrule
\multicolumn{10}{@{}l}{\textit{Continuous relaxation}} \\
\midrule
\texttt{GBDA}
  & \eopt{\lmopen} & \lmopen & \lmopen & \encopen & \lmopen & \lmopen & \clfopen & \nosup & \nosup \\
\texttt{PEZ}
  & \eopt{\lmopen} & \lmopen & \lmopen & \encopen & \lmopen & \lmopen & \clfopen & \nosup & \nosup \\
\midrule
\multicolumn{10}{@{}l}{\textit{Zeroth order}} \\
\midrule
\texttt{RandomSearch}
  & \eopt{\lmopen} & \lmopen & \lmopen & \encopen, \ecase{\encapi} & \nosup & \nosup & \clfopen & \lmopen & \lmopen, \lmapi \\
\texttt{BeamSearch}
  & \eopt{\lmopen} & \lmopen & \lmopen & \nosup & \nosup & \nosup & \nosup & \lmopen & \lmopen \\
\quad w/ aux.\ LLM
  & \eopt{\lmopen} & \lmopen & \lmopen & \encopen, \encapi & \nosup & \nosup & \clfopen & \lmopen & \lmopen, \lmapi \\
\texttt{RAL}
  & \eopt{\lmopen} & \lmopen & \lmopen & \encopen & \lmopen & \lmopen & \clfopen & \lmopen & \lmopen \\
\texttt{QCG}
  & \eopt{\lmopen} & \lmopen & \lmopen & \encopen, \encapi & \nosup & \nosup & \clfopen & \lmopen & \lmopen, \lmapi \\
\bottomrule
\end{tabular*}
\end{table}

\ifarxivsub\else
\clearpage
\input{neurips2026_template/checklist}
\fi

\end{document}